\documentclass[journal]{IEEEtran} 

\usepackage{siunitx} 
\usepackage{cite}
\usepackage{hyperref}
\usepackage{xspace}
\usepackage{url}
\usepackage{array}	
\usepackage[pdftex]{graphicx}
\usepackage{flushend}
\usepackage{balance}
\usepackage{orcidlink}

\begin{document}

\newcommand{\rhthree}{\textit{RBO~Hand~3}\xspace}
\newcommand{\refi}[1]{\textit{\nameref{#1}}}

\newcommand{\testcite}{ {\color{red}[CITE]}}
\newcommand{\TODO}[1]{{\textbf{\textcolor{blue}{TODO: }}}\textcolor{blue}{#1}}
\newcommand{\todo}[1]{\TODO{#1}}
\newcommand{\nth}[1]{}
\newcommand{\commentred}[1]{\textcolor{red}{#1}}
\newcommand{\steffen}[1]{\textcolor{violet}{[Steffen: #1]}}

\title{\rhthree \\ A Platform for Soft Dexterous Manipulation}

\author{Steffen~Puhlmann\orcidlink{0000-0002-3821-2127},~\IEEEmembership{Member,~IEEE}, Jason~Harris\orcidlink{0000-0001-8951-9877}, and~Oliver~Brock\orcidlink{0000-0002-3719-7754},~\IEEEmembership{Fellow,~IEEE}
\thanks{This work was funded by the European Commission (SOMA, H2020-ICT-645599), the Deutsche Forschungsgemeinschaft (DFG, German Research Foundation) under Germany’s Excellence Strategy - EXC 2002/1 ”Science of Intelligence” - project number 390523135 and German Priority Program DFG-SPP 2100 “Soft Material Robotic Systems” - project number 405033880.
All authors are with the Robotics and Biology Laboratory, Technische Universit\"at Berlin, Germany. }
}

\markboth{This paper is currently under revision in IEEE TRANSACTIONS ON ROBOTICS}%
{Shell \MakeLowercase{\textit{et al.}}: Bare Demo of IEEEtran.cls for Journals}
%

\maketitle


\begin{abstract}
We present the \rhthree, a highly capable and versatile anthropomorphic soft hand based on pneumatic actuation. 
The \rhthree is designed to enable dexterous manipulation, to facilitate transfer of insights about human dexterity, and to serve as a robust research platform for extensive real-world experiments. It achieves these design goals by combining many degrees of actuation with intrinsic compliance, replicating relevant functioning of the human hand, and by combining robust components in a modular design. 
The \rhthree possesses 16 independent degrees of actuation, implemented in a dexterous opposable thumb, two-chambered fingers, an actuated palm, and the ability to spread the fingers.
In this work, we derive the design objectives that are based on experimentation with the hand's predecessors, observations about human grasping, and insights about principles of dexterity. We explain in detail how the design features of the \rhthree achieve these goals and evaluate the hand by demonstrating its ability to achieve the highest possible score in the Kapandji test for thumb opposition, to realize all 33 grasp types of the comprehensive GRASP taxonomy, to replicate common human grasping strategies, and to perform dexterous in-hand manipulation. 
\end{abstract}

\begin{IEEEkeywords}
Robot hands, soft manipulation, dexterous manipulation, grasping, in-hand manipulation, soft robotics.
\end{IEEEkeywords}

%
\IEEEpeerreviewmaketitle

\section{Introduction}
\label{sec:introduction}

\IEEEPARstart{W}{e} present \rhthree, the third generation of soft robotic hands developed in our lab.  Similarly to the hands from the previous two generations, \rhthree is highly compliant, underactuated, pneumatically actuated, and fabricated predominantly from soft materials such as fabric or silicone rubber. In contrast to prior generations, however, the hand's dexterous abilities and its robustness for real-world experimentation have been substantially increased.

The first-generation hands were designed to take advantage of mechanical compliance~\cite{deimel2013compliant}. Compliance allows for safe and robust interactions because it dampens contact dynamics and results in large contact areas when the hand's morphology passively adapts to the shape of the environment.  Although the first-generation hands possessed only a single actuated degree of freedom, they were extremely successful in leveraging compliant interactions for robust grasping. This robustness exemplifies the substantial benefits of outsourcing aspects of perception and control to the compliant materials of the hand.

\begin{figure}[!t]
  \centering
  \includegraphics[width=0.9\linewidth]{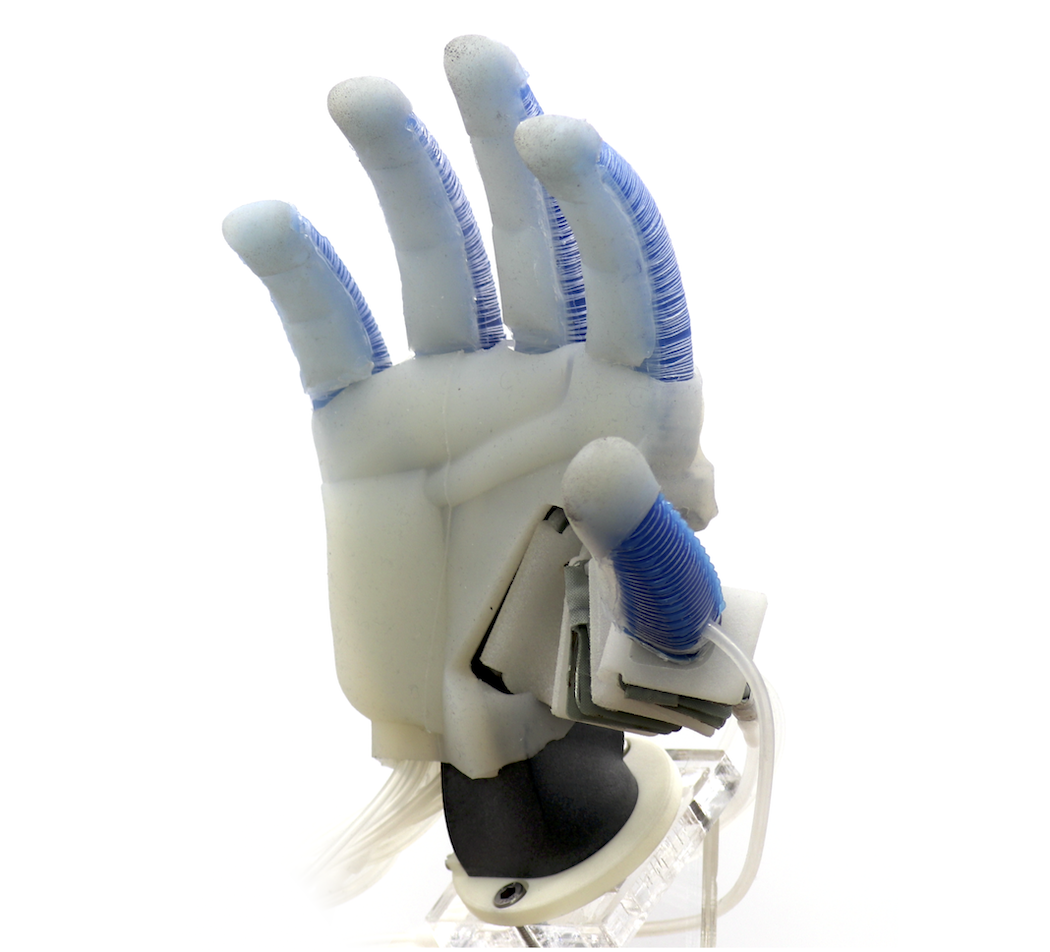}
  \caption{Anthropomorphic soft robotic \rhthree with 16 independent
    degrees of actuation and a dexterous, opposable thumb based on
    pneumatic actuation. The soft actuators are slightly
    inflated to realize a natural looking posture of a relaxed hand.
  }
  \label{fig:intro_hand}
\end{figure}

The second-generation hands featured seven actuated degrees of freedom, were anthropomorphic, and also fabricated from soft materials~\cite{deimel_novel_2016}. Thanks to its increased actuation abilities, the RBO~Hand~2 is able to reconfigure itself in many different ways. This ability results in a high level of dexterity, demonstrated by the hand's ability to replicate nearly the entire GRASP taxonomy~\cite{feix_comprehensive_2009}.  The ability to reconfigure itself also increases the variety of possible interactions between hand, object, and environment, including strategies such as sliding the object to the edge of the support surface or against a wall, before hand closure. We refer to this fruitful exploitation of features in the environment as the \textit{exploitation of environmental constraints}~\cite{eppner_exploitation_2015}. This exploitation of environmental constraints leads to improved robustness by compensating for uncertainty in sensing and control and facilitates successful grasping. In prior work, we found that this principle also forms the conceptual basis of human grasping~\cite{heinemann_taxonomy_2015,puhlmann_compact_2016}. 

The \rhthree, presented here, significantly extends the capabilities and features of the two previous generations. It exhibits a high level of versatility and robustness to support research in dexterous grasping and manipulation, including in-hand manipulation. The starting point for the design process of the \rhthree were three assumptions. First, we believe Mason's metaphor of a funnel, formulated in 1985, as "an operation that eliminates uncertainty mechanically"~\cite{mason1985mechanics}, to be the central enabling concept for dexterous manipulation. This metaphor provides a concise explanation for the effectiveness of exploiting environmental constraints, namely, reducing uncertainty through mechanical interactions. With \rhthree, we extend this funnel concept to dexterous manipulation in general by also considering exploitation of constraints that are provided by the manipulation platform. Second, we continue to rely on an anthropomorphic design, as the human hand is capable of producing the manipulation skills we would like to investigate. Also, much of the world around us is tailored to this design. Third, we believe it is important to develop the \rhthree as a research platform, i.e.~as a research tool that enables many hours of experimentation, without intermittent periods of complex repair. This ability is necessary to perform real-world experiments and to gather large amounts of real-world data required for learning-based approaches to manipulation.

In the following, we first motivate our design objectives in detail.  We then present how these objectives were translated into a specific mechanical design.  Subsequently, we evaluate the hand in its entirety, but also characterize selected aspects of the hand's design that represent reusable modules, suitable also for other applications. Our evaluation demonstrates that the \rhthree is a highly dexterous, capable, and robust hand. Future research using this platform will have to show conclusively whether our objective of producing a research platforms has been met successfully, but, over the last year, the \rhthree has served as a very effective and extremely reliable research platform for in-hand manipulation in our lab~\cite{Bhatt-2021}.

\section{Design Objectives} \label{sec:design_objectives}

In this section, we elaborate on the three main design objectives for the \rhthree 
and explain how we intend to achieve each of these objectives without going into specific implementation.

\subsection{Enabling Dexterous Manipulation}

Our experiences with the first two generation of soft robotic hands~\cite{deimel2013compliant, deimel_novel_2016, eppner2015planning} as well as insights obtained from human grasping experiments~\cite{heinemann_taxonomy_2015,puhlmann_compact_2016} both support our assumption that exploitation of constraints to compensate for uncertainty in sensing, modeling, and control plays a pivotal role in achieving robust manipulation. 

Mason's funnel metaphor~\cite{mason1985mechanics} is a standard explanation of these observations: During manipulation, deliberate contact with physical features produces physical constraints (e.g. table, wall, etc.)~\cite{eppner_exploitation_2015}. These physical constraints act as the metaphorical wall of the funnel. They guide, limit, and influence the manipulandum's state by realizing boundaries in some state dimensions, such as position or orientation, effectively reducing uncertainty. 
We argue that by purposefully constructing suitable manipulation funnels, the hand is capable of versatile and robust manipulation. 

However, physical constraints cannot only be found in the environment, but they can also be provided by the hand itself. The \rhthree can rearrange its physical features (e.g., fingers, palm, etc.), allowing it to produce a large variety of different funnels. 
The spatial arrangement of constraints can undergo changes through actuation as the manipulation progresses. This rearrangement can serve two purposes: first, reducing uncertainty by tightening the boundaries on the manipulandum's state,  or second, bringing the manipulandum into a desired state, either in preparation of the next manipulation step or to make progress towards the manipulation goal.

If dexterous manipulation is critically enabled by a hand's ability to produce suitable manipulation funnels, then the design of the hand must be capable of robustly producing and exploiting diverse arrangements of physical constraints. We now explain the requirements for this in more detail.

\begin{enumerate}
  \item \textit{Rearrangement:} To generate various manipulation
    funnels, the hand needs to be able to produce diverse spatial
    arrangements of physical constraints. At the basis of this lies
    the hand's ability to rearrange itself and to transition between
    many different postures.  The \rhthree must therefore possess a
    significant number of actuated degrees of freedom.
  \item \textit{Manipulation:} Changing the manipulandum's state
    mechanically requires the presence of forces.  These forces can be
    external, like gravity, or the result of rearranging the physical
    constraints.  Being able to exert various force patterns for many
    different hand postures facilitates diverse actuation of the
    manipulandum. This ability emphasizes the need for sufficient
    actuated degrees of freedom.
  \item \textit{Compensating for uncertainty:} The robustness of
    uncertainty-reducing funnels can be supported by inherent
    mechanical compliance. For example, when a soft hand's morphology
    handles complex contact dynamics so that they do not need to be
    addressed explicitly. Also, compliance allows the
    hand's posture and its morphology to passively adapt to the shape
    of the object or the environment, leading to larger contact areas
    and therefore to improved robustness in grasping and manipulation.
    However, compliance can also be
    detrimental~\cite{Ghazi-Zahedi-Deimel-Montufar-Wall-Brock-17-IROS}.
    To leverage the benefits of compliance while minimizing its
    drawbacks, the \rhthree must be able to adapt the direction of
    compliance.  Therefore, the hand must be inherently compliant,
    coupled with dexterity to modify this compliance through
    actuation.\\
\end{enumerate}

To achieve our design goal of producing a general platform for dexterous manipulation based on the idea of funnels, the \rhthree must be inherently compliant and possess a sufficient number of actuated degrees of freedom. This is required to enable the modification of compliance, the rearrangement of physical constraints, and the actuation of the manipulandum. Please note that these three different purposes of actuation will not be separate but overlap significantly during any real-world manipulation action.

\subsection{Leveraging Insights From Human Manipulation}

Anthropomorphism has shown to play a significant role for social human-robot interaction~\cite{anthropomorphism_hri}. However, we do not rely on a human-like design for social acceptance, but rather for taking inspiration from the result of millions of years of evolutionary optimization. 
Human manipulation capabilities remain substantially superior to those of robots. It seems therefore advantageous to facilitate the transfer of insights about human manipulation strategies to robot manipulation systems. This is facilitated by morphological and functional resemblance between the human and our robotic hand. It will therefore be an important design objective to \textit{replicate features of the human hand}---to the degree necessary to replicate the observed behaviors. 
Our starting point is thus an anthropomorphic design for the \rhthree. 

Choosing an anthropomorphic design also offers substantial guidance on how to select the relevant actuated degrees of freedom to satisfy the design objective from the previous section. We know that the human hand is capable of creating highly robust and versatile manipulation funnels. By mimicking its functionality, we hope to produce a manipulation platform that provides a substantial set of manipulation abilities. Our evaluation later in the paper confirms this conclusively.

\subsection{Support Extensive Real-World Experiments}

Research in real-world manipulation requires many hours of continuous, real-world experimentation. The nature of manipulation research, i.e.\ the need to repeatedly make and break contact with objects and the environment, imposes substantial requirements on the robustness of a useful manipulation platform. Considerations of robustness and minimization of downtime played a central role in the design of the \rhthree. We changed many features and manufacturing processes in this third generation to enable hundreds of hours of continuous grasping and manipulation experiments without failure.

An important decision for the design in this regard was motivated by the realization that a research lab cannot produce a complex artifact with the reliability of commercial, industrial products. Instead, we strived to maximize the robustness of all components as much as possible, while at the same time minimizing the complexity of repairs. As we will see, this design decision turns \rhthree into a capable and reliable research platform. 

\section{Related Work}

\begin{figure*}[!h]
\includegraphics[width=1.0\linewidth]{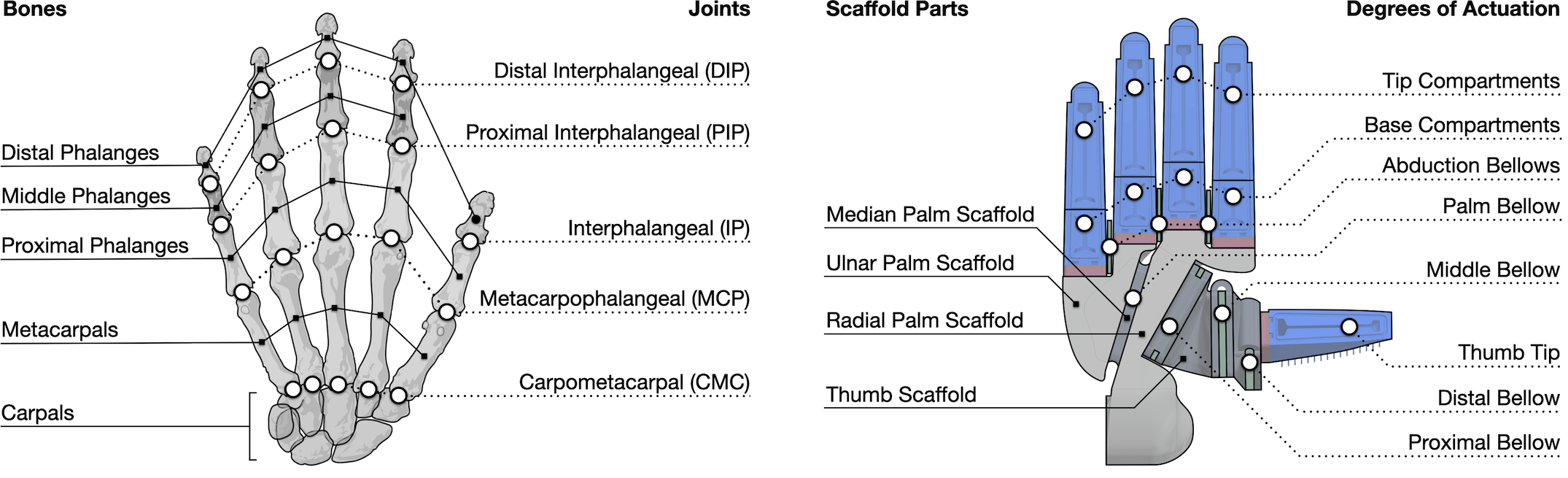}
\caption{The \rhthree is inspired by its human counterpart:
  nomenclature of joints and bones in the human hand~(left) and the
  corresponding naming of features in the \rhthree (right) }
\label{fig:comparison-human-rh3}
\end{figure*}

The space of possible robotic hands is huge, highlighted by the large variety of proposed designs~\cite{controzzi2014_hands_review, piazza_century_2019}. In this section, we discuss related works in the light of our design goals: enabling dexterous manipulation by combining many degrees of actuation with intrinsic compliance, leveraging transfer of human dexterity by realizing relevant human hand functionality, and supporting real-world experiments by exhibiting a high level of robustness.


Robotic grippers with a single or a few degrees of freedom based on servo motors and rigid links are by far the most common type of robotic hand, applied in various industrial applications~\cite{lundstrom1974industrial_grippers}. These hands are mostly tailored to solve specific tasks with high reliability, accuracy, and robustness. However, they do not support dexterous grasping and manipulation of many different objects in uncertain environments.


The advent of underactuated soft grippers~\cite{shintake2018_overview_softhands}  based on intrinsic compliance has shown improvements in grasping dexterity and robustness in the presence of uncertainty. These hands reduce control complexity to very few degrees of actuation while leveraging exploitation of environmental constraints and passive shape adaptation, effectively offloading aspects of sensing and control to the compliant hardware~\cite{laschi2014_soft_robotics_and_mc, muller2017_explanation_mc}. 

Various compliant actuation mechanisms have been proposed~\cite{zhou_soft_2015_different_gripping_mechanisms}.
Tendon-based soft grippers rely on highly compliant joints~\cite{dolla2010_sdm_hand, ma2013modular_open_source_SDM} or differrential tendon mechanics~\cite{friedl2018clash} to reliably grasp objects of unknown size and position.
Fin ray grippers~\cite{crooks2016_finray}, being derived from physiology of fish fins by incorporating crossbeams in their triangular fingers,  bend in the direction of contact, leading to significant shape adaptation. 
Grippers based on soft pneumatic actuators~\cite{walker2020_review_soft_pneumatic_actuators} have shown robust grasping behavior~\cite{hao2016universal_soft_gripper} also in extreme deep sea environments~\cite{galloway2016deep_reef_soft_gripper}.
Despite their reliable grasping capabilities, underactuated soft hands with few actuated degrees of freedom do not support dexterous manipulation or transfer of human strategies because of their inability to reconfigure themselves in many ways and because of their non-human design.

Anthropomorphic designs of underactuated soft hands achieve versatile grasping by suitably combining few independent degrees of actuation with passive shape adaptation based on intrinsic compliance~\cite{deimel_novel_2016, fras2018_anthropom_soft_hand, hussain2018_compliant_anthropomorphic_few_DOA, hundhausen2020_softhand}.
Some underactuated hand designs encode insights about synergistic actuation in the human hand in their compliant hardware~\cite{catalano2014_pisaiit_hand1, della2018_pisaiit_hand2}.
Although these hands are based on insights about human grasping behaviors and to some extend allow replicating human strategies, they lack the ability to reconfigure themselves to form diverse manipulation funnels and to exert many different force patterns which is required for dexterous in-hand manipulation.

Soft grippers that integrate multiple degrees of actuation are capable of executing various grasp types~\cite{teeple2020_two_comp_finger} and dexterous in-hand manipulation in the presence of uncertainty~\cite{abondance2020dexterous}.
Despite their ability to reliably grasp and manipulate various objects, direct transfer of human strategies is difficult due to high functional discrepancies to the human hand.


Highly dexterous capabilities have been demonstrated in rigid robotic hands by integrating many degrees of actuation in an anthropomorphic design~\cite{butterfass2001dlr_hand2, mouri2002_gifu_hand3, bae2012_allegro_hand}. 
Recently, the Shadow Dexterous Hand~\cite{kochan2005_shadow} demonstrated highly dexterous in-hand manipulation~\cite{openai_learning_2019} based on learning-based approaches, resulting in human-like behaviors.
Despite these impressive results, rigid hands lack intrinsic compliance, rely on complex mechanics, require exact modeling of contact dynamics, and often lack the required robustness for frequent and contact-intense interactions with the environment.

Integrating many actuated degrees of freedom with intrinsic mechanical compliance in an anthropomorphic design, allows highy dexterous and robust hands capable of reenacting human grasping behavior~\cite{schulz2001_new_ultralight_hand, wang2020_soft_hand, gilday2021_compl_hand_compl_wrist}.
In particular, the soft pneumatic BCL-26 hand~\cite{zhou2019_bcl26}, with 26 actuated degrees of freedom, has an actuated palm and an opposable thumb that achieves the highest possible score in the Kapandji test~\cite{kapandji1986clinical}. This hand exhibits a high level of versatility by achieving all grasp types in the GRASP taxonomy~\cite{feix_comprehensive_2009} and is capable of dexterous in-hand manipulation and in-hand writing.
Although the \rhthree possesses fewer degrees of actuation, our hand exhibits similar versatility and dexterity and also achieves the hightest possible scores in these tests. Furthermore, our hand more versatile, allowing it to grasp and manipulate a larger variety of objects, and we argue that the the functioning of \rhthree resembles more closely its human counterpart, allowing direct transfer of human strategies.
 




\section{Realization of the Design Objectives} \label{sec:realization_of_design_objectives}

We established three main design objectives for the \rhthree: enabling dexterous manipulation, facilitating the transfer of insights from  human manipulation strategies, and supporting extensive real-world grasping and manipulation experiments. To achieve these goals, we argued, the \rhthree needs to have an appropriate number of actuated degrees of freedom, possess inherent mechanical compliance, replicate important features of the human hand, and be designed to permit many hours of uninterrupted real-world experimentation. In the following sections, we elaborate on the the design choices we made to realize the design objectivse. 

The nomenclature for features of the human hand and the \rhthree is provided in Figure~\ref{fig:comparison-human-rh3}.
The manufacturing of the hand and its components is illustrated in Figure~\ref{fig:manufacturing}.

\subsection{Compliant Actuation} \label{sec:combining_actuation_and_compliance}

To combine mechanical compliance with many degrees of actuation, the \rhthree relies on pneumatic actuators based on soft materials, such as fabrics and silicone-rubber. 
These soft materials, together with compressible air, are intrinsically compliant. Although the soft actuators by themselves do not realize all of the design objectives we established above, they provide the building blocks for achieving these goals.
We now describe the actuators used in our hand design and explain their working mechanisms. Further below, we describe how the \rhthree integrates these actuators to realize the other design objectives.
\\

\subsubsection{PneuFlex Actuator} \label{subsec:pneuflex_actuator}

All five digits of the \rhthree rely on the soft pneumatic PneuFlex actuator~\cite{deimel2013compliant} which has been used already in our hand's predecessors. Because this actuator is an essential part of the design of the \rhthree, we will shortly reiterate its basic functioning: the PneuFlex actuator relies on an inflatable silicone air chamber whose radial expansion is constrained by a thread helix. The palmar side embeds a flexible but inextensible fabric which causes the actuator to bend upon inflation. 
The inflation profile, strength, and stiffness of a PneuFlex actuator can be easily adjusted through its geometry~\cite{deimel_novel_2016}, and as we will show below  (Fig.~\ref{fig:manufacturing}), manufacturing of this actuator is fast, simple, and low-cost. This allows for rapid prototyping and fast exploration of the design space which we extensively leveraged during the design process of the \rhthree. 
\\

\subsubsection{Bellow Actuator} \label{subsec:bellow_actuator}

Unlike its predecessors, the \rhthree relies on a second type of compliant actuator: the bellow actuator.
It realizes large rotation angles with a negligible bending radius based on flexible thermoplastic polyurethane (TPU)-coated nylon fabric. 
Its flat design allows stacking multiple bellow actuators to achieve movements in many different directions. This actuator therefore promotes the dexterity of the \rhthree by allowing complex kinematic structures with many compliant degrees of actuation in a small form factor. As we describe below, stacking bellows realizes the many degrees of actuation in the dexterous thumb.

A bellow consists of either a single or of multiple stacked inflatable pouches (Fig.~\ref{fig:thumb-actuators}).  When deflated, each pouch has a thickness of only \mbox{ca. 2 mm}. When placed between the two wings of a hinge joint, a bellow actuator realizes a rotational degree of actuation. The torque and opening angle of this joint increase with the level of inflation (Sec.~\refi{subsec:characterization_bellow_actuator}).

Manufacturing of our bellows (Fig.~\ref{fig:manufacturing}) is inspired by~\cite{yang2020layered} and based on heat-sealing the coated fabrics. However, our design permits stacking multiple bellows, because the pneumatic tubes enter the air chamber in the plane of the pouches. Also, bellow actuators that consist of multiple pouches require only a single tube, because their pouches are connected so that air can flow between their chambers.

Since the strength of a bellow actuator is proportional to the surface area of its air chamber, this type of actuator is not useful for all degrees of actuation in our hand design, for example for the fingers whose cross section area is constrained.

\subsubsection{Pneumatic Control} \label{subsec:pneumatic control}

As proposed in~\cite{deimel2016mass}, we control the air mass inside the pneumatic actuators of the hand. Air flow is regulated based on a linear forward model that takes input from air pressure sensors (Freescale MPX4250, 250kPa range, 1.4\% accuracy) for switching the valves (Matrix Series 320 - Model 321, max. 300 Hz). Controlling the mass instead of the pressure allows us to regulate the hand's preset behavior, i.e., the behavior in absence of contact. Therefore, the air masses and thus, the compliance of actuators does not change during contact-based deformation, allowing open-loop actuation through specification of desired compliance. This ability facilitates versatile behaviors that generalize across objects, as we will demonstrate below (Sec.~\ref{sec:evalulation_of_hand_design}).

\begin{figure}[!t]
  \centering
  \includegraphics[width=1.0\linewidth]{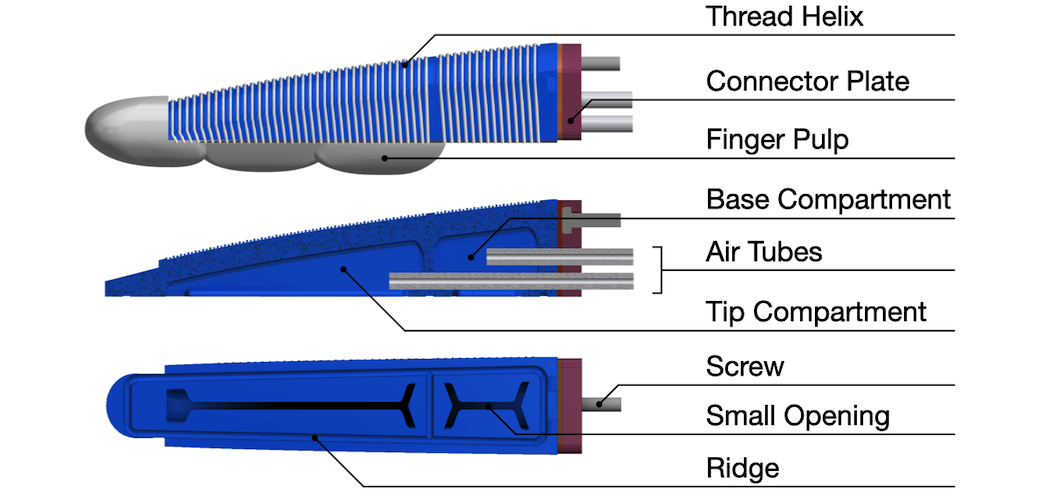}
  \caption{Two-Compartment Finger. Top: fully assembled finger; 
  	Center: cross-section, revealing the two air chambers and the tubing inside; 
  	Bottom: palmar side of the actuator with ridges and small
    openings for improved robustness.}
  \label{fig:two-comp-finger}
\end{figure}

\subsection{Enabling Dexterous Manipulation}

We argued that producing diverse manipulation funnels requires significant actuation. However, the number of pneumatic actuators --whose strength is proportional to their size at fixed pressure-- is limited by the outline of the hand. This limitation highlights the importance of choosing actuation that realizes relevant functionality. The \rhthree possesses 16 independent actuated degrees of freedom based on the soft actuators described above. Actuation of the \rhthree is inspired by the capabilities of its human counterpart. This allows us to address two design objectives at the same time: enabling dexterous manipulation by providing many degrees of actuation, and transfer of insights from human strategies by replicating relevant functionality of the human hand. We will now outline the related design features.
\\

\subsubsection{Two-Compartment Finger} \label{subsubsec:two-compartment-finger}

The four fingers of the \rhthree are based on the PneuFlex actuator with two actuated degrees of freedom (Fig.~\ref{fig:two-comp-finger}). 
The two compartments mimic the functionality of the human finger which can independently actuate its metacarpophalangeal (MCP) joint and its mechanically coupled proximal and distal interphalangeal (PIP and DIP, respectively) joints. This ability is crucial for performing the frequently observed precision grasp for which the fingertips move towards common point while MCP joints are flexed, and PIP and DIP joints extended. To adequately replicate this behavior in coordination with the kinematics of the actuated palm (which we describe below), we designed the little finger to be significantly shorter than the other fingers, as is the case for its human counterpart.

\begin{figure}[!t]
  \centering
  \includegraphics[width=0.9\linewidth]{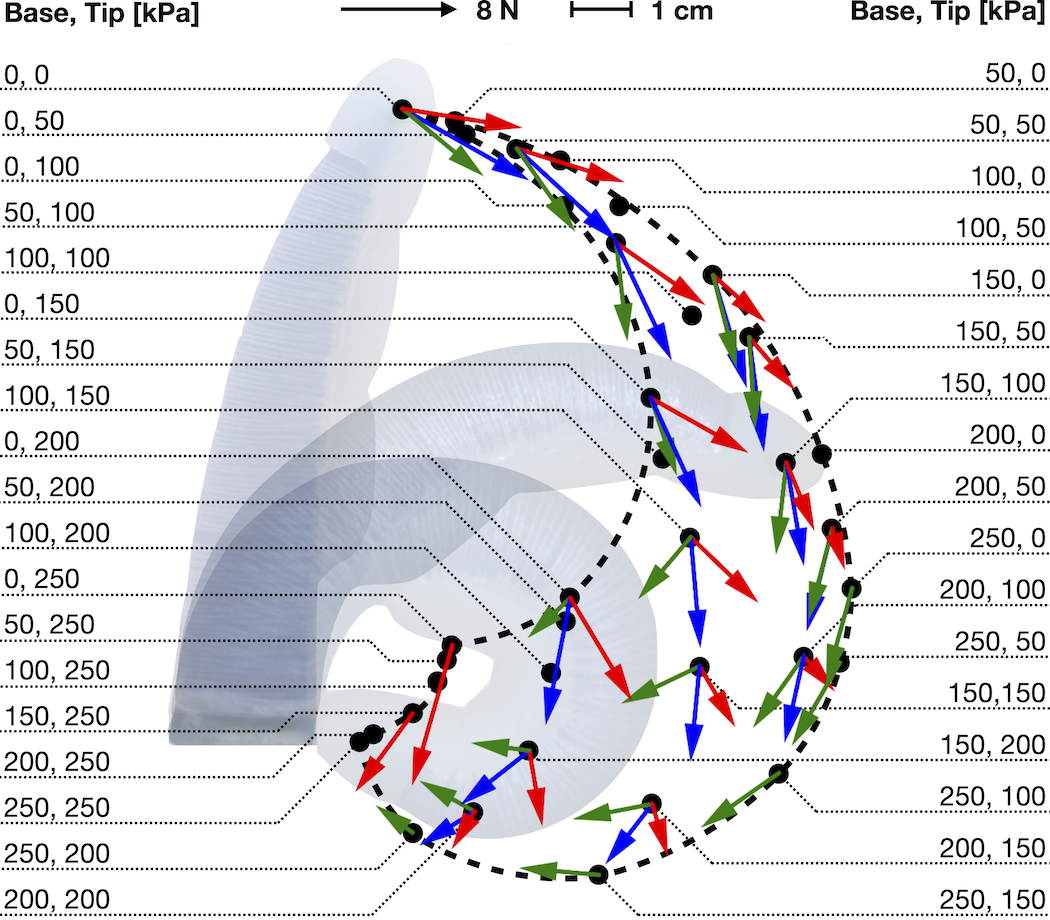}
  \caption{Work space (dashed line) and maximum forces (arrows) at sample positions inside the work space defined by inflation pressures of the two air chambers. Either only the base chamber (red), only the tip chamber (green), or both chambers (blue) are inflated to maximum pressure of 250 kPa. In case only a single chamber is maximally inflated, the other chamber remains at pre-defined pressure. Background shows three example finger postures (transparent) for 0,0 kPa (not inflated), 150, 100 kPa (partially inflated), and 250,250 kPa (maximally inflated).}
  \label{fig:two_comp_finger_experiment_results}
\end{figure}

Two independently actuated degrees of freedom allow the fingertip to reach a significant area on the finger's palmar side. 
This reachable workspace is illustrated in Figure~\ref{fig:two_comp_finger_experiment_results}. The figure also indicates the magnitude and direction of forces exertable at the fingertip in particular placements inside the workspace. Please note that the shape of the attainable workspace closely resembles that of the human finger, showing two arcs due to the independence of MCP and IP joints~\cite{finger_workspace}.

The finger is strongest when it is fully extended with a maximum force of ca. \SI{8.3}{N} and weakest when fully flexed with \SI{0}{N}. This is not surprising since the exerted force at a specific inflation level grows with the distance between the fingertip's actual position due to contact-based deformation and its corresponding position in absence of contact. In Section~\refi{subsec:characterization_two_compartment_finger}, we describe in detail how we obtained this data.

The significant size of the fingertip's workspace, together with the ability to exert significant forces over large regions of the workspace, illustrate that the new two-compartment finger contributes substantially to our design goal of producing diverse manipulation funnels and to vary the compliance of the fingers locally by leveraging contact interactions across multiple parts of the hand, mediated by the manipulandum. 

\subsubsection{Dexterous Opposable Thumb} \label{subsubsec:thumb}

The thumb design plays a pivotal role in achieving dexterity in the \rhthree. Our goal is to replicate the diverse abilities of the human thumb which is capable of the following movements: flexion moves the tip of the thumb in the direction of its pulp, perpendicular to the plane of the thumbnail.  Extension is the inverse movement to flexion.  Abduction moves the thumb away from the index finger. In the plane of the palm, this movement is also referred to as radial abduction and in the plane perpendicular to the palm, it is called palmar abduction.  Adduction is the opposite movement. Anteposition rotates the thumb towards to palmar side so that it points away from the hand perpendicular to the plane of the palm. Reposition does the opposite of anteposition.  Each of these motion pairs requires an actuated degree of freedom, thus further increasing the actuation within the \rhthree.

\begin{figure}[!t]
  \centering
  \includegraphics[width=1.0\linewidth]{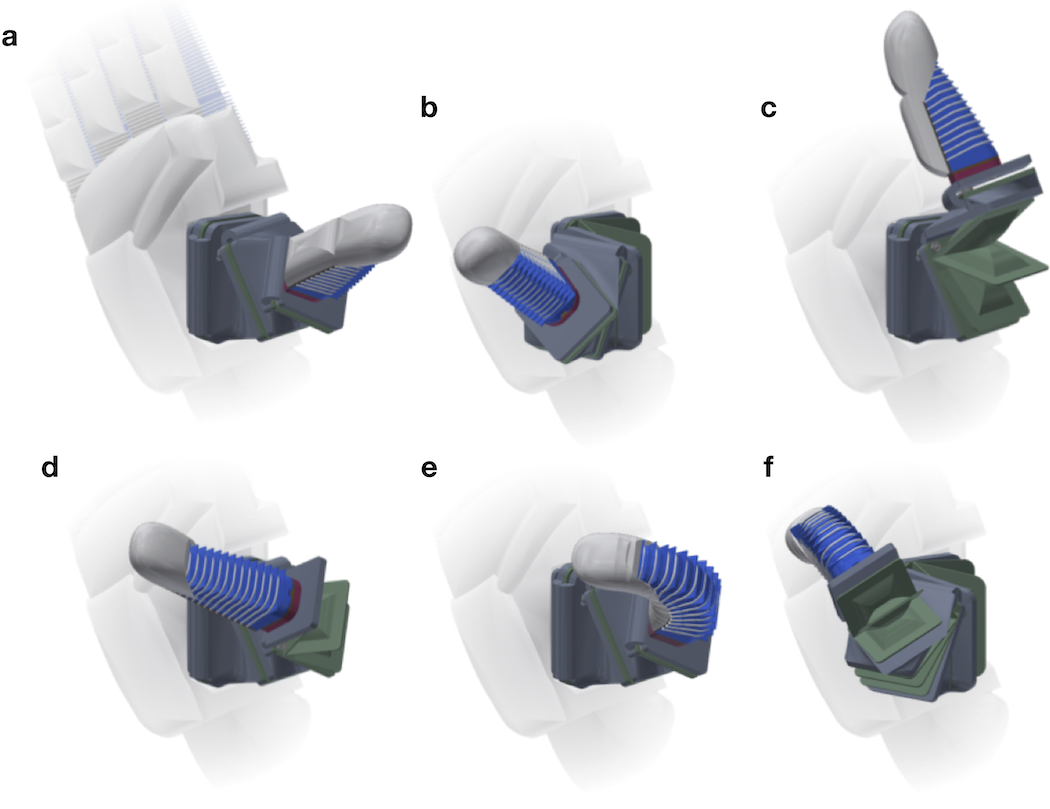}
  \caption{Thumb actuators and resulting movements upon inflation:
  \textbf{a)} no actuator inflated,
  \textbf{b)} Proximal Bellow for anteposition, 
  \textbf{c)} Middle Bellow for abduction, 
  \textbf{d)} Distal Bellow for flexion, 
  \textbf{e)} thumb tip for flexion,
  \textbf{f)} all actuators partially inflated.
  }
  \label{fig:thumb-actuators}
\end{figure}

The kinematic structure of the thumb of the \rhthree is inspired by its human counterpart. It possesses four degrees of actuation, realized by one single-chambered PneuFlex actuator for the tip of the thumb and by three bellow actuators for its carpometacarpal~(CMC) and MCP joints. 
Given the high strength of the bellow actuators (Sec.~\ref{subsec:bellow_actuator}), the design of the thumb's bellows was primarily guided by kinematic functionality, instead of specific force requirements. However, a stronger thumb could be achieved by increasing the size of its bellows, if desired. The overall design of the thumb and its movements are shown in Figure~\ref{fig:thumb-actuators}.

The thumb's four actuators are connected to a 3D-printed \textit{thumb scaffold} which is made of bendable TPU. 
The flexibility of this scaffold is modulated by the thickness of its material. Connecting two thicker, rigid plates with a thinner, more flexible sheet realizes a living hinge joint. The thumb scaffold constitutes a stack of three of these hinge joints that are actuated by bellows. Two bellow actuators (proximal and middle) mimic the movements of the human thumb's CMC joint. The proximal bellow is rotated towards the radial side by $30^\circ$ in the plane of the palm, relative to longitudinal axis of the fingers. Actuation of this bellow realizes an anteposition movement. The middle bellow is rotated in the plane of the palm so that the longitudinal axes of the fingers are perpendicular to the longitudinal axis of the thumb, and by  $90^\circ$ in the plane of its pouches towards the dorsal side. Actuating this joint realizes an abduction movement. The third bellow (distal) imitates the human thumb's MCP joint. It is rotated by $45^\circ$ about the thumb's longitudinal axis towards the palmar side and actuation of the distal bellow flexes the thumb at this joint. Finally, the human thumb's interphalangeal joint is represented as a short variant of the single-chambered PneuFlex actuator which forms the tip of the thumb. Actuating this joint flexes the thumb while bending the actuator. The position and orientation of the thumb's actuators was orchestrated to maximize its opposability which we evaluate in Section~\refi{subsec:kapandji}.

\begin{figure}[!t]
  \centering
  \includegraphics[width=1\linewidth]{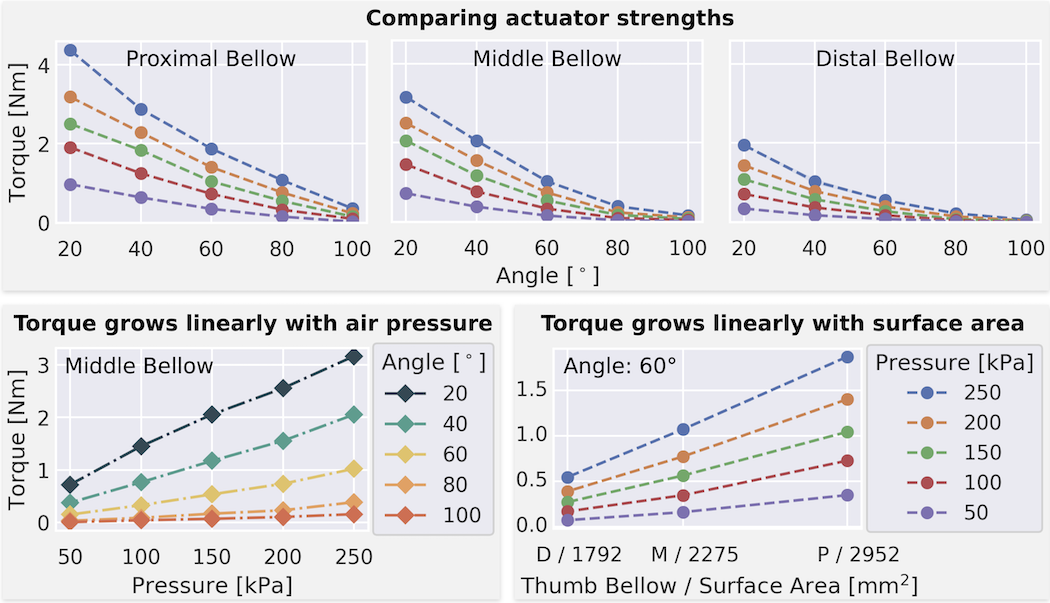}
  \caption{Torques achieved by the three bellow actuators of the thumb. Top row from left to right: torques exerted by the Proximal, Middle, and Distal Bellow for different opening angles and different inflation pressures. Bottom row: torque is pro\-por\-tion\-al to the inflation pressure (left) and to the surface area of the actuator's air chamber (right).
  }
  \label{fig:bellow_experiment_results}
\end{figure}

Integrating many degrees of actuation in a hand can lead to a high level of dexterity. However, to effectively manipulate many different objects, the hand also needs to be able to exert appropriate forces. 
The proximal, middle, and distal bellow actuators are therefore able to achieve high torques  (Fig.~\ref{fig:bellow_experiment_results}). Strongest torques for the Proximal, Middle, and Distal bellow were measured at an opening angle of $20^\circ$ (the experimental setup did not allow measuring smaller angles) with \SI{4.4}{Nm}, \SI{3.2}{Nm}, and \SI{1.9}{Nm}, respectively when maximally inflated to \SI{250}{kPa}. Their strengths decreases with the opening angle. We explain in detail in Section~\refi{subsec:characterization_bellow_actuator} how we obtained these measurements.
Integrating many strong degrees of actuation in the highly dexterous opposable thumb contributes significantly to the hands ability to reconfigure itself and to exert many different force patterns for providing diverse manipulation funnels and manipulating a large variety of objects.

As we explain below, also other design features of the \rhthree benefit from strong bellows, including palm hollowing and finger spreading.
\\

\subsubsection{Palm Hollowing}

During our design studies, we learned that the commitment to an anthropomorphic hand design also necessitates the implementation of palm hollowing~\cite{kapandji1971physiology} via an additional actuated degree of freedom in the \rhthree.  

In the human hand, the palm is spanned by the carpal and metacarpal bones (Fig.~\ref{fig:comparison-human-rh3}).   
In contrast to the index and the middle finger whose CMC joints permit only very limited movement, the ring and the little finger can flex at their metacarpal bones at their respective CMC joints. The CMC joint of the little finger is the most mobile of the four fingers and allows flexion of up to $30^\circ$~\cite{hamill2006biomechanical}. Flexion at these joints improves opposition of the ring finger and especially of the little finger with respect to the thumb. It also enables the palm to better adapt its shape to objects or to the environment. Additionally, the described flexion at the CMC joints contributes to the aforementioned inwards movement of the fingertips during the frequently observed precision grasp.

To imitate the functioning of these CMC joints, the palm of the \rhthree consists of the \textit{radial palm scaffold}, representing the carpal and metacarpal bones of the index and the middle finger, and of the \textit{ulnar palm scaffold}, representing the corresponding bones of the ring and the little finger. These two parts are connected via the \textit{median palm scaffold} which houses the \textit{palm bellow} actuator (Fig.~\ref{fig:comparison-human-rh3}).  Actuating the palm bellow results in a motion that imitates simultaneous flexion at the CMC joints of the little finger and ring finger of the human hand. As mentioned above, the kinematics of the palm was coordinated with the length of the little finger. 

As we will demonstrate in Section~\refi{subsec:kapandji}, this additional degree of actuation greatly improves thumb opposition and thereby improves the hand's dexterity and versatility.
\\

\subsubsection{Finger Abduction}

The fingers of the human hand are able to move apart and together (abduction and adduction, respectively), thanks to their condyloid type metacarpophalangeal joints. 
This permits the fingers to better encompass the object and to exert forces from different directions which facilitates robust grasping and manipulation. 
To replicate these movements with the \rhthree, we place \textit{abduction bellows} (each consists of two pouches) between the base-compartments of neighboring fingers (Fig.~\ref{fig:comparison-human-rh3}) which deform laterally when inflating these actuators. 
The ability of abducting the fingers increases their workspaces and thus improves the dexterity and versatility of the \rhthree. 
\\

\subsubsection{Soft Layer}

The \rhthree exhibits substantial inherent mechanical compliance. This newest version of our soft hands further increases the compliance of its fingers and its palm by equipping them with a soft layer. While details of this soft layer, such as palm ridges, are purely cosmetic, its overall shape and material are again inspired by functionality of the human hand.

For imitating the fleshy mass of human finger pulps, the palmar sides of the fingers and of the thumb are covered with a thick (up to \SI{10}{mm}) layer of silicone material. This material is much softer (shore hardness~00-30) than the silicone of the pneumatic actuators (shore hardness~A-30). The palm is covered by a glove that is also made of the softer silicone. These soft layers adapt to the shape of an object, resulting in larger contact areas and thus, to improved robustness in grasping and manipulation. Additionally, the finger pulps significantly reduce the cavity when fingers are fully flexed, which for previous iterations of the RBO Hand caused the object to slip in some grasp postures~\cite{deimel_novel_2016}. 
At the same time, the harder silicone of the actuators maintains structural integrity of the fingers and withstands high air pressures. In combination, the fingers benefit form both types of material by exhibiting a high degree of compliance and improved strength at the same time.
\\

\subsubsection{Summary: Enabling Dexterous Manipulation}

In total, the \rhthree has 16 actuated degrees of freedom based on intrinsically compliant pneumatic actuators: eight in the four fingers, four in the thumb, three for abduction of the fingers, and one for palm hollowing. These degrees of actuation are sufficient to replicate relevant functioning of the human hand, as described above. For improved compliance, the palm and the palmar sides of the fingers are covered by a soft layer for passive shape adaptation.
We will show in Section~\refi{sec:evalulation_of_hand_design} that with these design features, the \rhthree is highly dexterous, capable of producing a wide range of manipulation funnels and thus achieving our design objective.

\subsection{Support Extensive Real-World Experiments}

In this section, we will elaborate on some of the design and manufacturing details that have contributed to making the \rhthree a robust experimental platform, capable of operating for hundreds of hours without hardware failure.
\\

\subsubsection{Modularity}

To serve as a versatile research platform, the \rhthree is based on a modular design. This design is a key strategy of achieving the robustness required for a research platform. Modularity reduces the number of distinct parts and, in our design, greatly facilitates repair simply by replacement.

The radial palm scaffold, which can be mounted to a robot arm via a dovetail mount, constitutes the base of the modular hand design. The fingers are connected with screws to the ulnar and radial scaffold via custom-built \textit{connector plates}~(Fig.~\ref{fig:manufacturing}). To reduce space and avoid clutter, air tubes of the fingers are guided by tunnels through these scaffolds. The pouches of the bellow actuators are also connected via screws to their respective scaffold: the pouches of the proximal, middle and distal bellow are screwed to the thumb scaffold and the palm bellow is screwed to the median palm scaffold (Fig.~\ref{fig:comparison-human-rh3}).  The four scaffolds of the \rhthree are connected to each other also via screws. The soft silicone glove can be put on and taken off easily, as humans do with common gloves. It contains pockets between the fingers which house the bellow actuators for finger abduction.

The modularity of the \rhthree allows for quick and easy replacement of broken parts and for effective exploration of hand designs when physically evaluating newly developed hand parts or combinations of different variants of the digits.
\\

\subsubsection{Improved Robustness}

\begin{figure*}
\includegraphics[width=1.0\linewidth]{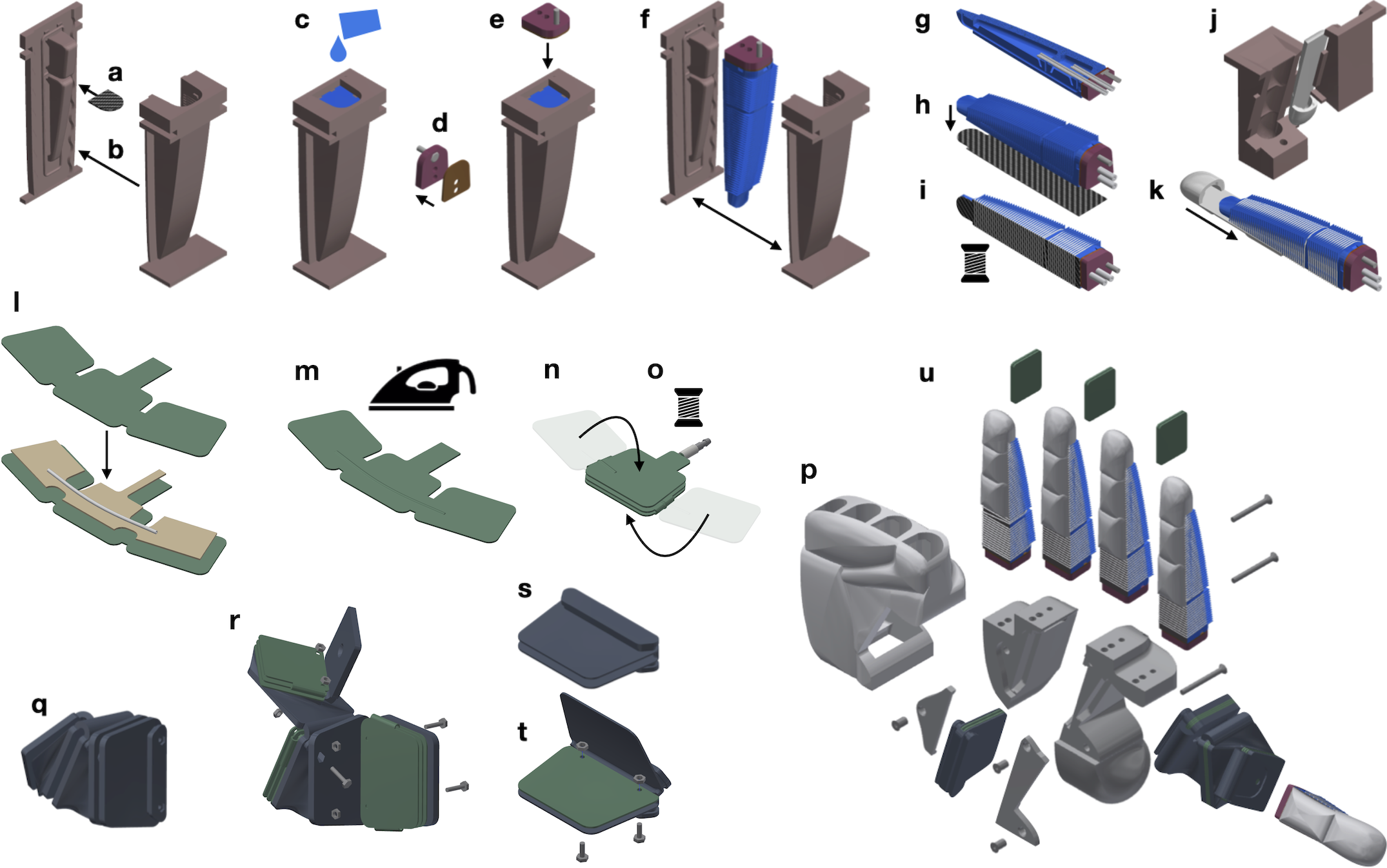}
\caption{
%
Manufacturing of the \rhthree. 
A complete hand can be built within five days, including curing time. Material cost is less than US\$ 250 with 3D printing as the highest cost factor.
%
\textbf{Manufacturing of the Two-Compartment Finger:} 
\textbf{a)}~Prior molding, a piece of inextensive PET-based fabric (black and white striped) is placed inside the 3D-printed mold (grey-brown) to reinforce the wall between the two compartments of the actuator.
\textbf{b)}~Top and bottom part of the mold are connected tightly. 
\textbf{c)}~Silicone (blue) of type Dragon Skin\texttrademark~10 (Smooth-On) is poured into the mold. 
\textbf{d)}~Connector plate is built from a laser-cut and -engraved piece of acrylic glass (violet). A hexagonal engraving holds a screw (grey) in place, which later connects the finger to the scaffold. The connector plate is finished by glueing a piece of woolly fabric (brown) to its backside. 
\textbf{e)}~Connector plate is placed inside the mold with the woolly fabric facing downwards to soak in the wet silicone.
\textbf{f)}~After curing, the actuator is robustly connected to the acrylic connector plate. Both are removed from the mold.
\textbf{g)}~Cross-section of the actuator. Two silicone tubes (grey) are guided through designated holes in the connector plate and punctures in the silicone material towards the two compartments. The tube of the tip compartment passes through the base-compartment and the reinforced wall close to the palmer side of the finger.  Holes and punctures are sealed with Sil-Poxy\texttrademark~(Smooth-On) silicone adhesive.
\textbf{h)}~The bottom side of the actuator is sealed by attaching the \textit{passive layer}, a sheet of inextensive PET-based fabric soaked with wet Dragon Skin\texttrademark~10 silicone (black and white striped).
\textbf{i)}~After the passive layer cured, a helix structure of nylon thread (white) is spun around the actuator.  
\textbf{j)}~Soft finger pulp (grey) is cast in a separate mold, using Ecoflex\texttrademark~0030 (Smooth-On) silicone.
\textbf{k)}~Finally, the finger pulp is glued to the actuator with Sil-Poxy\texttrademark. 
%
%
\textbf{Manufacturing of the Thumb Tip:}~The tip of the thumb has only a single compartment and is manufactured by following 
steps~\textbf{b)}~to~\textbf{k)}. However, the hexagonal laser-engraving of the connector plate houses a nut instead of a screw. The tube is inserted through a puncture on the dorsal side of the actuator instead of a hole in the connector plate. 
%
%
\textbf{Manufacturing of the Bellow Actuator:} (exemplary for proximal bellow)
\textbf{l)}~TPU-coated nylon fabric (green) and a baking paper (beige) are laser-cut into shape. 
Baking paper is placed between two precisely stacked sheets of nylon fabric whose coated sides face each other. 
A silicone tube~(light grey) of \SI{1.5}{mm} diameter is placed between baking paper and nylon fabric to ensure air flow between neighboring pouches. 
%
\textbf{m)}~The fabric pieces are heat-sealed using a steam iron for approximately one minute with 220$^\circ$C.
The backing paper prevents the TPU coating from melting together which results in an air chamber. 
\textbf{n)}~The actuator is folded at the connections of neighboring pouches to realize a stack.
%
\textbf{o)}~ The actuator connects to a tube via a plastic hose fitting inserted into its outlet. A piece of silicone tube around the actuator-facing side of the fitting serves as rubber seal. Finally, air tightness is ensured by tightly spinning a nylon thread around the outlet at the location of the rubber seal.
\textbf{p)} The soft silicone-based glove is molded separately. 
%
\textbf{Assembly of the Hand:}
\textbf{q)}~The thumb scaffold (anthracite) is 3D-printed using TPU plastic. It houses three bellow actuators with differently shaped pouches.
\textbf{r)}~The flaps of the thumb scaffold are bent open. The respective bellows are attached with screws and nuts to designated holes in each of the hinge joints.
\textbf{s)}~The median palm scaffold is also 3D printed using TPU plastic.
\textbf{t)}~A bellow consisting of a single pouch that has the same shape as the pouches of the priximal thumb bellow is connected with screws and nuts to the median palm scaffold.
\textbf{u)}~The fingers are connected to the 3D printed radial palm scaffold and ulnar palm scaffold using the screws inside the connector plates of the fingers and nuts. The tubes are guided by tunnels through the scaffolds.
The thumb tip is connected to the thumb scaffold with a screw and the nut inside the connector plate of the thumb tip.
The thumb scaffold and the median palm scaffold are placed in cut-outs of the ulnar and the radial scaffold. They are fixated by two plates which are connected with screws and sleeve nuts. The silicone glove is put on. 
Finally, the abduction bellows are placed inside pockets of the silicone glove between the fingers. 
\vspace{2cm}
}
\label{fig:manufacturing}
\end{figure*}

To enable the \rhthree to perform in real-life tasks, we improved the longevity and durability of its actuators by modifying their design and manufacturing (Fig.~\ref{fig:manufacturing}). First, we increased the thickness of the silicone walls to prevent ruptures and penetrations.  This modification also brings about stronger fingers by withstanding higher pressures. Second, we updated the molds for casting the top-piece of the actuators so that the resulting cast has \mbox{i) significantly} smaller openings at the palmar side, resulting in a larger area of adhesion with the \textit{passive layer} (the inextensible layer), and  \mbox{ii) ridges} at its bottom side so that the adhesion bond better withstands shear forces which occur during inflation (Fig.~\ref{fig:two-comp-finger}). Third, we increased the number of turns of the thread helix, effectively reducing the radial force each turn exerts onto the actuator's surface, which otherwise may cut the silicone material after repeated inflation with strong air pressures.

All of these modifications have substantially increased the robustness an reliability of the two-compartment fingers. After implementing these changes and while writing this paper, we have not experienced a sufficient number of failures to provide a reliable mean time to failure. We estimate this time to be at least 300~hours of \textit{continuous} use (more than one month of eight-hour days).

\begin{figure}[!t]
  \centering
  \includegraphics[width=1\linewidth]{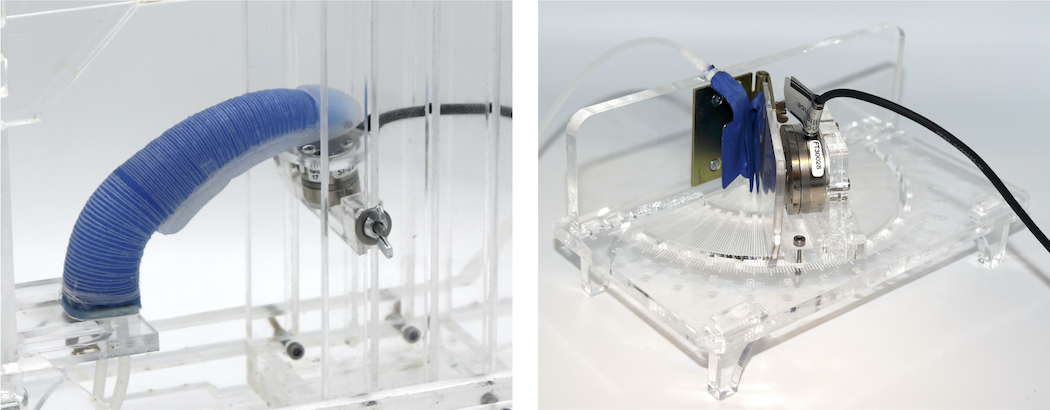}
  \caption{Custom-built experimental setups for actuator characterization.
    Left: Setup for measuring work space and forces of a
    two-compartment finger. The finger is fixated to the setup at its
    base. A force-torque sensor whose position and orientation are adjusted 
    to be in front of the fingertip measures exerted forces when the finger
    is actuated.  Right: Setup for measuring torques of a bellow
    actuator. Bellow is placed between the wings of a hinge
    joint. One wings is fixated while the other can rotate. 
    Actuation increases the hinge joint's opening angle. 
    A force-torque sensor whose contact surface is aligned with
    the rotating wing measures the exerted forces at radius of
    \SI{5}{cm}.  }
  \label{fig:characterization_setup}
\end{figure}

\section{Actuator Characterization}

In this section, we characterize the actuators of the \rhthree. In particular, we analyze the work space of the two-compartment actuator and the forces it can exert at the fingertip. We then analyze the maximum torques realized by the bellow actuators.
While we control the hand's behavior based on air mass (Sec.~\ref{subsec:pneumatic control}), we characterize its actuators by controlling the air pressure. We do this because air pressure, which in contrast to air mass changes in the presence of external forces, is better suited for specifying maximum inflation states.

\subsection{Characterization of the Two-Compartment Finger} \label{subsec:characterization_two_compartment_finger}

In Section~\refi{subsubsec:two-compartment-finger}, we demonstrated the two-compartment finger's large workspace and its ability to exert strong forces over large regions of its workspace, highlighting its significance for the \rhthree to form diverse manipulation funnels.

To determine the quantitative results shown in Figure~\ref{fig:two_comp_finger_experiment_results}, we mount the finger at its base to a custom-built experimental setup (Fig.~\ref{fig:characterization_setup}). This setup contains a force-torque sensor which can be freely positioned and oriented relative to the finger. We sample multiple fingertip positions by inflating the two air chambers separately to pre-defined air pressures, ranging between \SI{0}{kPa} and \SI{250}{kPa}, in intervals of \SI{50}{kPa}, resulting in 36 pressure combinations. To record the fingertip position and the maximum forces at this position, we first inflate the two air chambers to one of the pre-defined air pressures. We then measure the distance between the tip of the two-compartment actuator and the base of the finger along the horizontal and vertical axes to obtain the fingertip position. In a next step, we adjust the pose of the force-torque sensor to be positioned directly in front of the fingertip while its contact surface is perpendicular to the direction of flexion. To infer the direction of measured forces, we determine the orientation angle of the force-torque sensor inside the plane of flexion.

We then inflate 1)~only the base chamber, 2)~only the tip chamber, and 3)~both air chambers to the maximum air pressure of \SI{250}{kPa}. When only a single air chamber is maximally inflated, the other chamber remains at its pre-defined pressure. When the finger reaches maximum inflation, we record the exerted forces with the force-torque sensor.  Since inflation of base and tip chamber can result in different directions of flexion, the pose of the force-torque sensor needs to adjusted for the three cases separately. For each of the pre-defined pressure combinations we repeat the three cases of maximum inflation for five times.

Based on the orientation angle and the measurements of the force-torque sensor, we determine the average direction and intensity of exerted forces for the different fingertip positions and the different cases of maximum inflation (Fig.~\ref{fig:two_comp_finger_experiment_results}). On average, the standard deviation of the five repetitions per pressure combination is only ca. \SI{0.06}{N}.

\nth{say that this was not the little finger but one of the other three}
\nth{better comparison with human finger's workspace}

\subsection{Characterization of the Bellow Actuator} \label{subsec:characterization_bellow_actuator}

%
In Section~\refi{subsubsec:thumb}, we demonstrated the strengths of the bellow actuators by showing that they achieve high torques which is required for exerting appropriate force patterns to effectively manipulate various objects.

To determine the exerted torques of a bellow actuator~(Fig.~\ref{fig:bellow_experiment_results}), we mount it to a custom-built characterization setup~(Fig.~\ref{fig:characterization_setup}). This setup, inspired by~\cite{sun2013characterization}, consists of two acrylic plates which realize the two wings of a hinge joint. While the position and orientation of one of these wings is fixed, the other can rotate around the hinge joint's rotational axis. We place the bellow actuator between the two wings so that the opening angle of the hinge joint increases upon inflation. In addition, the setup contains a force-torque sensor whose position and orientation can be adjusted so that its contact surface is aligned with the rotating wing and the center of its contact surface is kept at a radius of \SI{5}{cm}.

For measuring the torques at a specific opening angle, the force-torque sensor is positioned so that the freely moving wing cannot exceed this angle. We then inflate the actuator to pre-defined air pressures that range between \SI{50}{kPa} and \SI{250}{kPa} in intervals of \SI{50}{kPa}, resulting in five possible inflation states. The force-torque sensor then measures the exerted force and the torque is obtained by multiplying this force by the radius. We repeat this process five times for different opening angles which ranges between $20^\circ$ and $100^\circ$ in intervals of $20^\circ$. 
We performed this experiment for each of the three bellow actuators of the thumb.

We computed the average torque for each bellow actuator, inflation pattern, and opening angle (Fig.~\ref{fig:bellow_experiment_results}).
Our measurements indicate that the torque of a bellow actuator is proportional to the air pressure and to the area of the actuator's air chamber, which is in agreement with the physical principle that force equals pressure times surface area. The average standard deviation of the exerted torque across the five repetitions per inflation pressure and opening angle is less than \SI{0.1}{Nm}. 

%

\section{Evaluation of the \rhthree} \label{sec:evalulation_of_hand_design} 

In this section, we evaluate the \rhthree and demonstrate that the specific design choices indeed realize the design objectives.  We demonstrate that the \rhthree is capable of highly dexterous and versatile manipulation behavior. Similar to prior work~\cite{Bhatt-2021}, we use a mixing board to regulate the inflation mass inside the hand's actuators, making it easy for novice operators to replicate the behaviors presented here. Furthermore, we also report on our experiences with using the \rhthree as a platform for manipulation research.

\subsection{Thumb Opposability} \label{subsec:kapandji}

\begin{figure}[!t]
  \centering
  \includegraphics[width=1.0\linewidth]{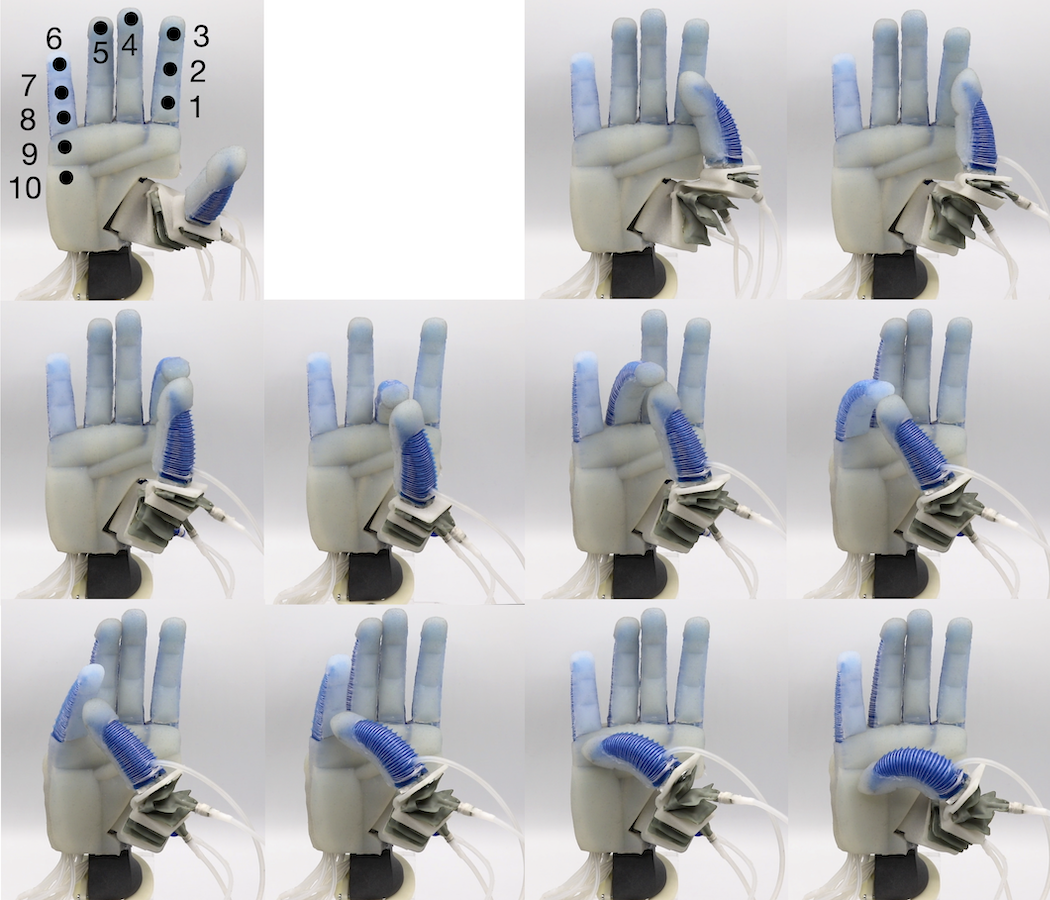}
  \caption{
  The RBO Hand 3 achieves the highest possible score in the Kapandji test thanks to its dexterous, opposable thumb which is able to reach all ten locations on the hand. 
  }
  \label{fig:kapandji}
\end{figure}

We use the Kapandji test~\cite{kapandji1986clinical} to compare the functionality of our thumb design to its human counterpart while evaluating its opposability.
This test was originally developed to evaluate hand motor functions of patients after stroke or surgery. In the robotics research community, this test has established itself as an informative tool for evaluating and comparing functionality of thumb designs. 
For the Kapandji test, the tip of the thumb has to touch ten different locations on the hand (Fig.~\ref{fig:kapandji}). The score is determined by the number of locations that the thumb is able to touch, with a zero score indicating no thumb opposability and a score of ten indicating maximum opposability. 

To perform the Kapandji test with the \rhthree, we pre-recorded different hand poses in which the thumb touches one of the desired hand locations and replayed these poses in the right order. The \rhthree achieves the highest possible score, as shown in Figure~\ref{fig:kapandji}. This is achieved by not only inflating the actuators of the thumb and the fingers, but also by actuating the palm bellow actuator which rotates the ring and the little finger towards the thumb. Including this palm bellow is necessary for reaching points five to ten, highlighting the significance of this degree of actuation. 

The results of the Kapandji test demonstrate the dexterity of the thumb design and its significance in the hand's ability to create diverse manipulation funnels.

\begin{figure}[!t]
  \centering
  \includegraphics[width=0.6\linewidth]{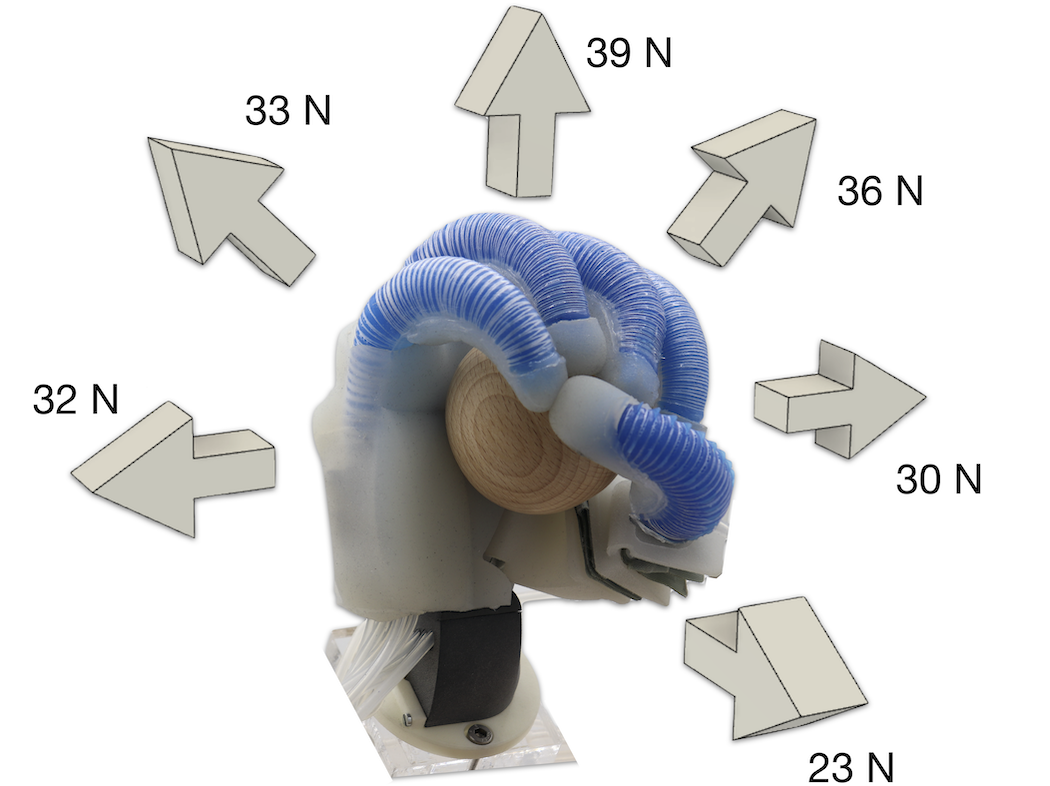}
  \caption{Object pull-out experiment.  A force torque sensor connected the object via an inextensible wire measures the required force to pull the object out of the closed hand in different directions. Arrows indicate pulling directions and numbers indicate corresponding pull-out force, averaged over five trials. Standard deviation is below \SI{3}{N} for each direction.}
  \label{fig:object_pullout}
\end{figure}

\subsection{Grasp Postures} \label{subsec:grasp_postures}

\begin{figure*}[!t]
  \centering
  \includegraphics[width=0.8\linewidth]{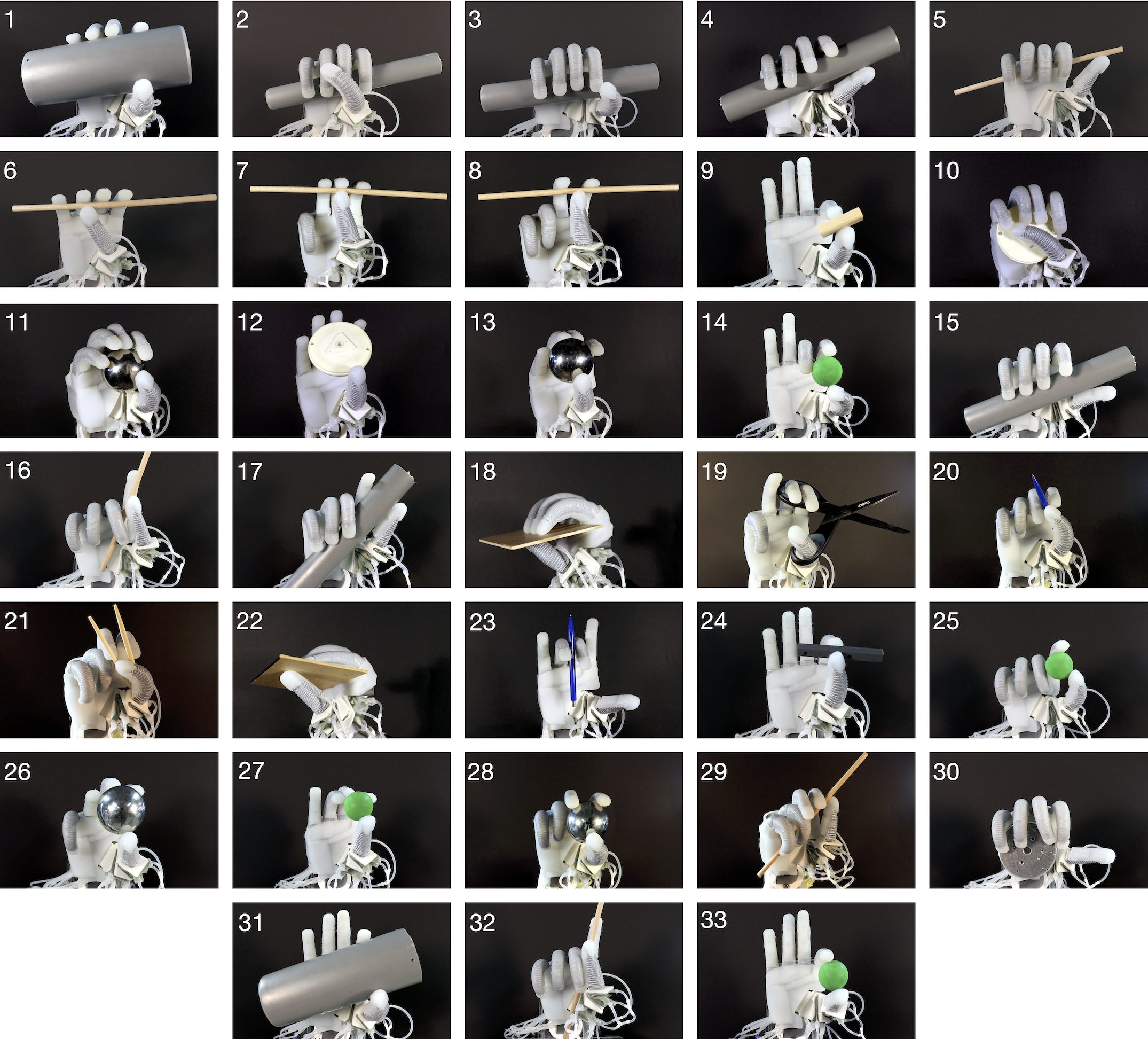}
  \caption{The \rhthree is able to replicate all 33 grasp postures of the most comprehensive used GRASP taxonomy.}
  \label{fig:feix_taxonomy}
\end{figure*}

We assess the dexterity and versatility of the \rhthree by showing that it is capable of achieving many different grasping postures. A common practice to measure a hand's grasping capabilities is to reenact common human grasps. The GRASP taxonomy~\cite{feix_comprehensive_2009} is the most comprehensive and well-established taxonomy to date. It encompasses the 33~most commonly observed grasp types in humans with 17~different object shapes.  We argue that a hand's ability to achieve many different grasp postures is also a good indicator for its ability to provide various spatial arrangements of physical constraints and to exert many different force patterns. 

To reproduce grasps, we pre-record air mass-based actuation patterns for each of the grasp types separately.  During hand closure, an operator holds the object in an appropriate position while the hand replays the actuation pattern. A grasp is successful if the hand holds the object for at least 10~seconds. We repeat this procedure three times per grasp type.

The \rhthree is able to perform all 33 grasps repeatedly, with three successful consecutive trials (Fig.~\ref{fig:feix_taxonomy}), highlighting the dexterity and versatility of our hand design. This ability is achieved by integrating many degrees of actuation that functionally replicate the human hand with compliance for passive shape adaptation.

The predecessor of the \rhthree achieved only 31 grasps, failing the light tool grasp~(5) and the distal type grasp~(19), because the fully flexed fingers formed a too large a cavity on the palmar side of the fingers. The \rhthree reduces this cavity through its soft finger pulps.  Furthermore, the increased dexterity of the thumb enables the scissors in the distal type grasp. Subjectively, the \rhthree performs both the Kapandji test and the GRASP taxonomy with motions that appear much more human-like.

\nth{We would have better arguments for the contribution of compliance to solving this task, if we would have used only a hand full of closing synergies that - thanks to compliance - achieve all 33 grasps of the GRASP taxonomy.}

\subsection{Grasp Strength} \label{subsec:grasp_strength}

We demonstrate the overall strength of the \rhthree by showing that it can firmly hold an object onto which pulling forces are applied. The ability to withstand external forces, such as gravity, depends on the strength of the hand's actuators and indicates its ability to grasp and manipulate heavy objects. The higher the required forces to pull-out an object, the stronger the hand's actuators and vice versa.

We measure the required pulling force in six different directions (Fig.~\ref{fig:object_pullout}). For this, a wooden sphere of \SI{6}{cm} diameter is placed inside the hand.  A small metal hook on the object's surface connects it to a force-torque sensor via an inextensible wire. This hook always points into the pulling direction. Hand closure is realized by inflating the actuators with pre-determined air masses, resulting in a firm power grasp. We then manually pull the force-torque sensor away from the hand in one of the six pulling directions. The force-torque sensor measures the forces exerted on the object. We gradually increase the pulling force until the object is released. We repeat this procedure five times for each pulling direction.

For each pulling direction, we compute the average force required to release the object. Highest force is required for the distal direction (parallel to an extended finger) with \SI{39}{N} and the lowest force in palmar direction (orthogonal to the plane of the palm) with \SI{23}{N}. This is not surprising since in distal direction, the four fingers combine their strengths by forming a barrier and in the palmar direction, only the fingertips and the thumb tip prevent the object from being pulled out. In radial and ulnar direction (parallel to the extended thumb), no digits are available that could form a barrier. However, the strong thumb and the actuated palm compensate for this lack so that mediocre forces of \SI{30}{N} and \SI{32}{N} are required for pull-out, respectively. The standard deviation of the required force less than \SI{3}{N} in each pulling direction.

The results of this experiment show that the \rhthree is capable of firm grasps while significantly exceeding maximum pull-out forces of comparable other pneumatic hands, such as the BCL-26 Hand with {21.9}{N}~\cite{zhou2019_bcl26}. The forces achieved by our hand are sufficient for experimentation of manipulation with a large variety of everyday objects that do not exceed our hand's maximum payload of ca. \SI{3.9}{kg}. They also demonstrate that soft pneumatic actuators made of silicone and fabric materials are capable of producing large forces. Taken together with the beneficial friction properties of silicone rubber, grasp strength and robustness is competitive with those of  ``hard'' anthropomorphic hands, such as the Shadow Dexterous Hand with a maximum payload of \SI{4}{kg}~\cite{shadow_website}.

\subsection{Replicating Human Grasping Strategies} \label{subsec:replicate_grasping}

\begin{figure}[!t]
  \centering
  \includegraphics[width=1\linewidth]{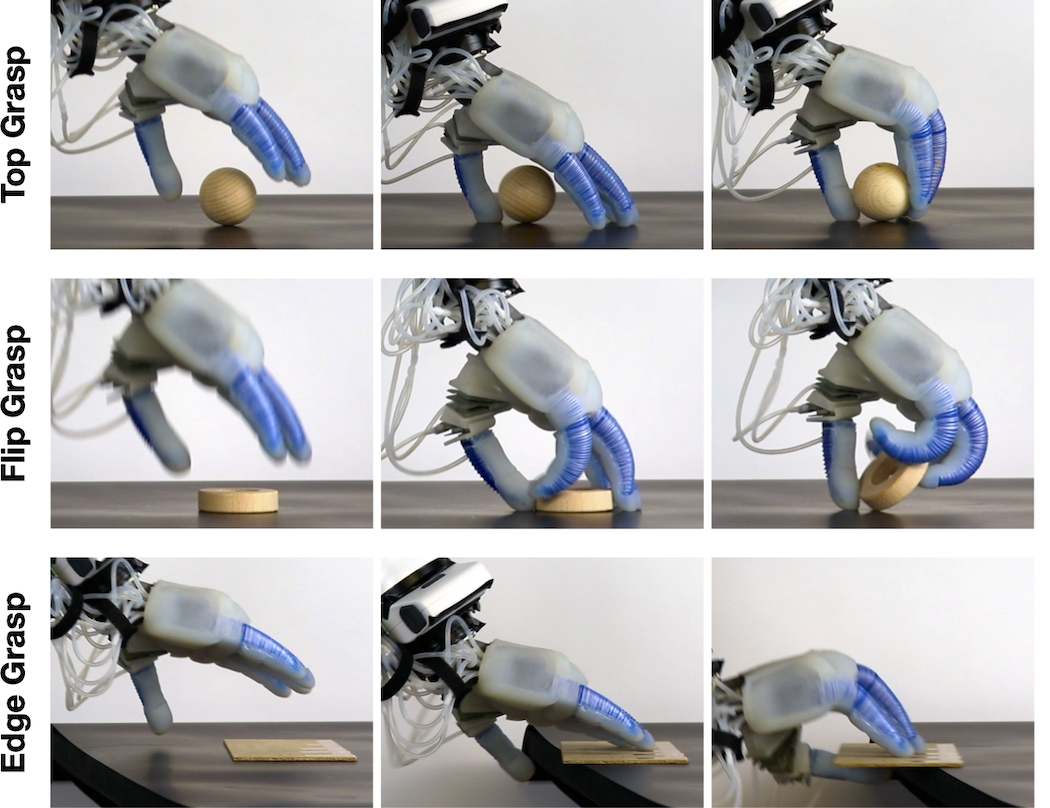}
  \caption{\rhthree replicates the three most commonly observed human grasping strategies. 
  Top grasp: during hand closure, fingertips move inwards while being guided by the support surface. Flip grasp: the thumb fixates the object before it is rotated by the fingertips. Edge grasp: hand slides the object towards the edge of the support surface before grasping it from the side.
  From left to right: Hand approaches the object, exploitation of environmental constraints (the tabletop), hand closes to grasp the object.}
  \label{fig:replicate_human_grasping}
\end{figure}

To verify the design objective that the \rhthree should facilitate transfer of human strategies, we replicate the three most frequent table-top grasping strategies observed in humans~\cite{heinemann_taxonomy_2015, puhlmann_compact_2016}. We refer to these strategies as top grasp, flip grasp, and edge grasp (Fig.~\ref{fig:replicate_human_grasping}).  In more than 78\% of the 3400~trials that were conducted in our prior work, participants performed one of these three strategies. 

When performing the top grasp, the fingertips touch the support surface which guides their inwards movement until a precision grasp is established.  For the flip grasp, the precision grasp is preceded by a manipulation maneuver during which the object is fixated at one of its sides with the thumb while the fingers lift the other side to reveal a large contact surface. During the edge grasp which was observed most frequently for flat objects, the hand makes extensive use of the tabletop while sliding the object toward the edge of the support surface to reveal the object's bottom side, before grasping it with a precision grasp.

For each of the three grasp types, we pre-recorded closing synergies defined by sequences of air-masses and joint trajectories of the robotic arm (Panda by Franka Emika). We then placed objects on the tabletop and replayed these synergies in open-loop control. The same grasping synergy was executed during each execution of a specific grasp type. Because of its compliance, the fingers and the thumb of the \rhthree passively adapt their shape to the environment and to the object, allowing it to reliably grasping a large variety of objects with these three pre-defined motions (see supplement video). 

The \rhthree is able to reliably replicate the three grasping strategies, closely resembling the behavior of its human counterpart for the top grasp and the edge grasp. For the flip grasp, however, the \rhthree grasps objects with the back of its fingers, which is atypical in humans. 
This behavior follows from the high fingertip forces required for this particular strategy, which our hand realizes by simultaneously inflating the two compartments of its fingers to high inflation levels. Nevertheless, the \rhthree still follows the same underlying strategy as humans when revealing a large contact surface by rotating the object about an axis formed by the thumb. 

These experiments demonstrate that the \rhthree is indeed capable of replicating relevant functionality of its human counterpart. The ability to grasp many different objects with the same grasping synergy also highlights the significant contribution of compliance to dexterity and versatility by reducing uncertainty and allowing robust constraint exploitation. This provides additional support for our design objectives discussed earlier.

\subsection{In-Hand Manipulation} \label{subsec:IHM}

\begin{figure}[ht!]
  \centering
  \includegraphics[width=1\linewidth]{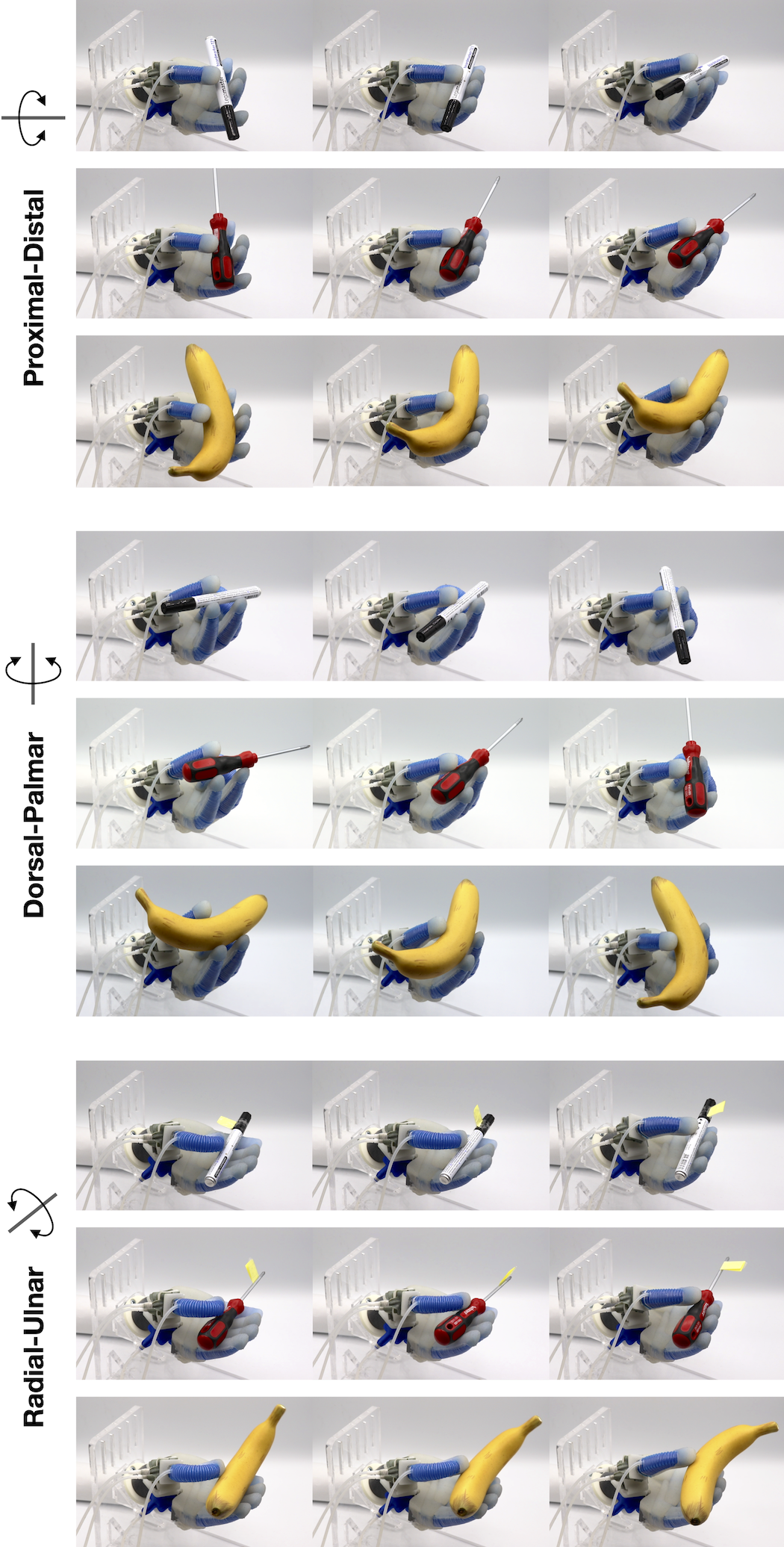}
  \caption{The \rhthree performs simple in-hand manipulations. Three different objects (pen, screw driver, and plastic banana) are rotated inside the hand about the proximal-distal axis~(top), dorsal-palmar axis~(middle), and radial-ulnar axis~(bottom). For each of these three rotation types, the hand performs the same actuation pattern irrespective of the object. To achieve these maneuvers, only a subset of the actuators are inflated while the rest of the hand serves as a spatial arrangement of compliant constraints which passively adapt their shape to the movement of the object, resulting in complex behavior despite simple control.}
  \label{fig:in_hand_manipulation}
\end{figure}

We demonstrate the dexterity of the \rhthree by performing three simple in-hand manipulations during which the hand rotates three different objects (pen, screw driver, and plastic banana) about the proximal-distal axis, the ulnar-radial axis, and the palmar-dorsal axis (Fig.~\ref{fig:in_hand_manipulation}). 
The ability to perform diverse manipulations with a wide variety of objects inside the hand shows a high level of dexterity.

For each of the three manipulations, we manually place objects inside the hand and replay pre-recorded actuation trajectories based on air masses. These actuation trajectories are the same for all objects. During manipulation, only a subset of the hand's actuators is inflated while the others passively and continuously adapt their shape to the movement ob the object. Specifically, for the proximal-distal rotation, only the little and the index finger are actuated. For the dorsal-palmar rotation, only index and middle finger, and for the radial-ulnar rotation, only the tip of the thumb, the middle bellow, and the base-compartments of the four fingers are actuated. In each of these cases, the rest of the hand that is in contact with the object serves as a spatial arrangement of constraints that --thanks to compliance-- passively rearranges itself while maintaining contact with the object and firmly holding it in place. The fact that the same actuation trajectories result in successful manipulations for three objects that significantly differ in size, shape, and weight, further highlights the profound contribution of compliance to dexterity and versatility. 

While these experiments highlight the robustness of our hand's soft manipulation abilities against variation in object properties, we showed in prior work that this robustness also applies to execution speed and initial hand-object configuration. The latter allows the \rhthree to repeat the same in-hand manipulation up to 140 times, before the object falls out of the hand~\cite{Bhatt-2021}

We argue that the presented manipulations constitute robust manipulation funnels, because the configuration of object and constraints imposed by the hand change over time due to actuation-based forces which reduces uncertainty in the object position and orientation.
Although the reconfiguration of constraints is rooted in actuation, most of the hand motion---and therefore most of this rearrangement---is caused by passive, intrinsically compliant constraints that exert forces by trying to retain their original pose. Uncertainty is reduced continuously because the compliant constraints support the hand to firmly holding the object. We therefore view these manipulations as robust funnels based on purposeful combination of actuation and compliance, providing further evidence for the generality of the \rhthree and further support for our design objectives.

\subsection{\rhthree as a Platform} \label{subsec:research_platform}

A suitable research platform for dexterous manipulation should exhibit dexterity and robustness. So far, in this section, we have demonstrated the dexterous capabilities of the \rhthree. We now want to report on our experience with the \rhthree as the main research platform for manipulation in our lab.  As we reported before, we are currently not able to provide a detailed quantitative analysis of mean time to failure, as we have not encountered a sufficient number of failures. But based on our experience during in-hand manipulation experiments in our lab~\cite{Bhatt-2021}, we estimate the two-compartment fingers to provide at least 300 hours of continuous use, whereas the pouch actuators provide at least 60 hours of continuous use. We believe these numbers to be impressive for a non-commercial hand. However, they are probably insufficient for a productized version of the \rhthree. Therefore, we pursued the strategy of making it very easy to repair the hand.

The first repair strategy is simple replacement of either a two-compartment finger or a pouch actuator. Other parts have not failed yet. Replacing either of them does not require any specific skills. It can be achieved with two Allen wrenches and a regular wrench. Replacing the finger takes ten minutes and replacing a pouch takes five minutes, once the operation has been performed once or twice. This means that in all observed failures, the downtime is at most ten minutes.

There are three main failures we have observed. First, the helical thread surrounding the fingers can get displaced from strong inflation, causing small bulges in the finger.  Second, a finger can get damaged at its based and become leaky. Third, the seam of the pouches can break and become leaky. 

The finger bulge and the leaky pouch can be repaired, even by a layperson, in minutes, by using a common iron and Sil-Poxy silicone adhesive. To fix the finger bulge, the affected part of the finger is covered with Sil-Poxy; this takes about three minutes. Sil-Poxy has to cure for eight hours but then the finger is read for use again. A leaky pouch can be fixed by ironing the leaky seam; this also takes about three minutes.

By significantly increasing the robustness of its components and by making these components very easy and quick to repair, we have developed a research platform for dexterous manipulation with unprecedented availability in the context of academic research. 

We tested the  actuator design during frequent use of the \rhthree during which its bellows were inflated repeatedly.
During continuous use, the bellow actuator reliably withstands air pressures of up to \SI{300}{kPa}. The bellow actuator fails more quickly than the two-compartment finger. But, as with the two-compartment finger, we have not experienced a sufficient number of failures to provide a reliable mean time to failure. We estimate this time to be at least 60~hours (more than one week of eight-hour days) of \textit{continuous} use.

\section{Limitations and Future Work} \label{sec:limitations_and_future_work}

In this section we elaborate on the limitations of the \rhthree and point to future work that is
currently being developed in our lab.

\subsection{The Ambivalence of Soft Material Robotics}

We demonstrated that deliberate exploitation of intrinsic mechanical compliance significantly improves robustness, facilitating dexterous grasping and manipulation. However, soft material robots are often believed to be limited to low forces, they pose new challenges for sensorization, exhibit imprecise actuation, and it is difficult to find accurate analytic models. We now want to discuss these drawbacks in more detail.

Maximum forces achieved by soft pneumatic fingers tend to be lower than that of their rigid counterparts, because flexional forces can be diverted when soft fingers bend away from the object due to low lateral and torsional stiffness. Furthermore, soft material actuators can break at high levels of air pressure which are necessary for achieving high forces. 
These problems can be addressed by embedding a rigid skeleton into the soft finger in order to achieve kinematic stiffness and transmission of forces while decoupling contact location and acting forces~\cite{lotfiani2020torsional}. Also, changing the geometry of the PneuFlex actuators allows modulating the finger's stiffness, force, and bending profiles, for example through thicker walls that withstand higher air pressures.

Sensorization of compliant hands is challenging since sensors need to be based on flexible materials to provide sensing abilities without detrimental effects on compliance. This introduces significant constraints on possible sensor designs which often rely on complex fabrication or costly components. As we will discuss below, we are currently working on novel soft sensor technologies that have very little effect on the compliance or design space of soft actuators.

Actuation of the \rhthree is less accurate than that of rigid hands, because the PneuFlex actuators can exhibit different behaviors, due to manufacturing-based differences. Furthermore, small errors in the pressure-based estimates of air masses inside the pneumatic actuators accumulate over repeated inflation cycles, making it difficult to precisely repeat and predict hand movements over long time periods. Thus, tasks that require high levels of precision and repeatability are beyond the scope of this hand.

Finally, efficiently and accurately modeling the behavior of soft hands interacting with the environment, especially the deformation of soft materials, is difficult and a topic of ongoing research. The lack of accurate models renders traditional approaches to grasping and manipulation inapplicable to soft-material robot hands.

We argue, however, that many of the perceived shortcomings of soft material robotics disappear when suitable control and planning methods are used. The results presented here and in prior work~\cite{Bhatt-2021} demonstrate that a high level of robustness, not precision in the sense of accuracy and repeatability, is key to successful manipulation. We showed that exploitation of physical constraints and compliance are important principles for achieving robustness and versatility in manipulation. When these principles form the basis of grasping and manipulation approaches, a different kind of precision becomes relevant.  It is not achieved through accurate models and precise control but instead results directly from physical, compliant interactions~\cite{Bhatt-2021}.

\subsection{Sensorization of the \rhthree}

Our hand demonstrates impressive grasping and manipulation despite pure open-loop control.
Of course, dexterity and versatility could be further improved by incorporating sensory feedback, enabling responsive behaviors for autonomous grasping and manipulation tasks. We are therefore working on feedback control based on pressure sensing, and are also developing various other soft sensor technologies in our lab. 

These developments include liquid-metal strain sensing for propriozeption. Combining several of these sensors allows also inferring contact on the entire actuator, including forces, and contact location~\cite{wall2017method, wall2019multi}. 
We also investigate soft tactile sensing based on piezoresistive fabric and a flexible printed circuit board to realize multiple tactile units in a flat and compliant design~\cite{pannen2021low}.
Furthermore, we are developing an acoustic sensor, relying on sound propagating through the soft finger to infer contact location, contact intensity, and contacted material~\cite{zoller2018acoustic, zoller2020active}. 
 
These sensor designs have demonstrated their abilities when attached to the PneuFlex actuators in isolation. In future work, we will investigate how feedback from these sensors can further improve the dexterity when integrated in the \rhthree, for example by adding responsiveness to the aforementioned manipulation funnels.  
We believe that with these novel soft sensor technologies, we can address the criticism of mechanical compliance regarding sensorization of soft hands.

\subsection{Pneumatic Control}

For controlling air mass trajectories inside the actuators, pneumatic valves repeatedly open and close for a specific time duration. Since these valves currently support only a binary state of being either fully open or fully closed, hand movements can exhibit tremor due to rapid changes in the air flow when the states of valves change.  
Although we found these oscillations to have only minor effect on manipulation performance, we will update the control of the \rhthree to rely on proportional valves, providing continuous opening states. In combination with precise mass flow sensors, this updated control scheme will significantly improve smoothness and accuracy of hand movements.

\section{Conclusion}

We presented the \rhthree, an anthropomorphic soft robotic hand with 16 independent degrees of actuation that exhibits a high degree of versatility and dexterty. 
The hand enables dexterous manipulation, supports transfer of human strategies, and serves as a reliable research platform for contact-intense manipulation experiments 
Following these design objectives, the \rhthree integrates many degrees of actuation with compliance in a robust, anthropomorphic design that replicates relevant functioning of the human hand. The hand is built in a modular fashion from low-cost and easily accessible materials which allows rapid prototyping for exploring of the design space for future advancements.

The \rhthree achieves the highest possible score in the Kapandij test by combining the dexterity of its truly opposable thumb with an actuated palm. The hand can perform all 33 grasp postures from the most comprehensive GRASP taxonomy, highlighting the ability of the hand to reconfigure itself an various ways. This facilitates versatility and robustness, because it allows the hand to form diverse manipulation funnels. We demonstrated the hand's strength in an object-pullout experiment in which it can hold an object in the presence of external forces of up to \SI{39}{N}.

We also showed that the \rhthree is able to replicate the most commonly observed grasping strategies in human single-object tabletop trials which heavily rely on exploitation of environmental constraints. This ability highlights the robustness and durability of the hand which is required for safe interactions with the environment in the presence of uncertainty. It also shows the hand's ability to functionally replicate its human counterpart. 

Furthermore, we conducted in-hand manipulation experiments and demonstrated the hand's ability to rotate different objects inside the hand through reconfiguration of compliant, actuated constraints.  The fact that the hand can grasp and manipulate a variety of objects, despite performing the same actuation pattern, demonstrates that intrinsic compliance facilitates robustness, versatility, and dexterity by absorbing contact dynamics and by allowing the hand to passively adapt to the shape of the environment and the object. Purposefully leveraging compliance effectively reduces uncertainty by outsourcing aspects of sensing and control the compliant hardware so that complex behaviors result from simple control which facilitate successful grasping and manipulation.

Finally, we discussed limitations of our hand design and outlined future work, including sensorization and improved pneumatic actuation.

\bibliographystyle{IEEEtran}
\bibliography{references}

\vspace{1.0 cm}

\begin{IEEEbiography}[{\includegraphics[width=1in,height=1.25in,clip,keepaspectratio]{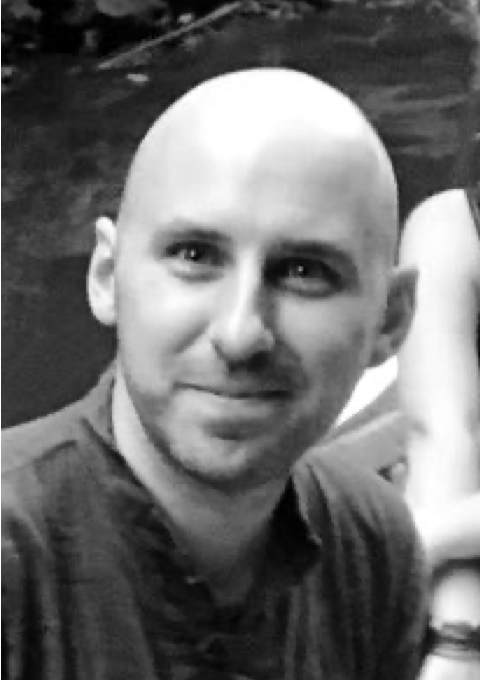}}]{Steffen Puhlmann}
is a Ph.D. candidate at the Robotics and Biology Laboratory (RBO) at the School of Electrical Engineering and Computer Science at Technische Universit\"at Berlin, Germany. He received his Bachelor's degree in Computer Science from Freie Universit\"at Berlin in 2014 and his Master's degree in the Computer Science track Science of Intelligence from Technische Universit\"at Berlin in 2017. His research interests include analysis of human grasping and manipulation strategies, design and development of anthropomorphic soft robotic hands, soft sensing, and dexterous in-hand manipulation. 
\end{IEEEbiography}

\begin{IEEEbiography}[{\includegraphics[width=1in,height=1.25in,clip,keepaspectratio]{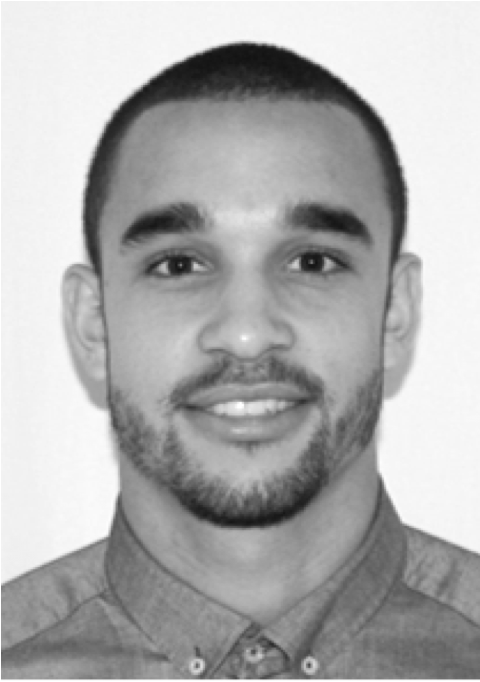}}]{Jason Harris}
pursues his Master's degree in Computer Science at Technische Universit\"at Berlin, Germany.  He received his Bachelor's degree in Mechatronic Systems in 2017 at the University of Applied Sciences Brandenburg, Germany in combination with a dual studies program at the Siemens Technical Academy. From 2018 to 2020, Jason Harris worked as a student research assistant at the Robotics and Biology Laboratory (RBO) at Technische Universit\"at Berlin. Since 2021 he is working as a robotics engineer at Gestalt Robotics GmbH in Berlin. His research interests include robotic manipulation, design and manufacturing of robotic hands, robotic motion and task planning, and robust behavior in robotics.
\end{IEEEbiography}

\begin{IEEEbiography}[{\includegraphics[width=1in,height=1.25in,clip,keepaspectratio]{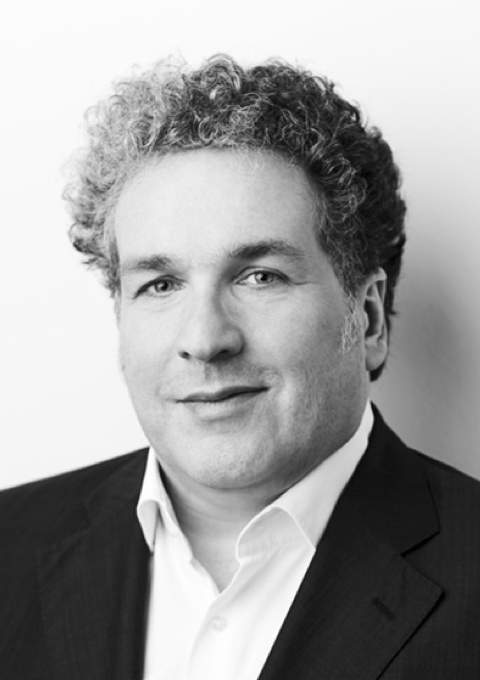}}]{Oliver Brock} is the Alexander-von-Humboldt Professor of Robotics in the School of Electrical Engineering and Computer Science at Technische Universit\"at Berlin, a German "University of Excellence". He received his Ph.D. from Stanford University in 2000 and held postdoctoral positions at Rice University and Stanford University. He was an Assistant and Associate Professor in the Department of Computer Science at the University of Massachusetts Amherst before moving back to Berlin in 2009. The research of Brock's lab, the Robotics and Biology Laboratory (RBO), focuses on robot intelligence, mobile manipulation, interactive perception, grasping, manipulation, soft material robotics, interactive machine learning, deep learning, motion generation, and the application of algorithms and concepts from robotics to computational problems in structural molecular biology. Oliver Brock directs the Research Center of Excellence "Science of Intelligence". He is an IEEE Fellow and was president of the Robotics: Science and Systems Foundation from 2012 until 2019.
\end{IEEEbiography}

\end{document}